\documentclass{article}

\usepackage{microtype}
\usepackage{graphicx}
\usepackage{subfigure}
\usepackage{booktabs} 
\usepackage{amsmath}
\usepackage{array} 

\usepackage{multirow}
\usepackage{adjustbox}
\usepackage{amsmath,amssymb}
\usepackage{xspace}
\usepackage{xcolor}
\usepackage{hyperref}
\usepackage[most]{tcolorbox}

\usepackage{hyperref}

\usepackage[accepted]{mlsys2025}

\mlsystitlerunning{Where Is VLA Inference Time Spent?}

\newcommand{\na}{--}
\begin{document}

\twocolumn[
\mlsystitle{Where Is VLA Inference Time Spent? Bottlenecks and Optimization Interactions Across GPUs}



\mlsyssetsymbol{equal}{*}

\begin{mlsysauthorlist}
\mlsysauthor{Muhammad Danish Waseem}{equal,to}
\mlsysauthor{Ahmed Ali-Eldin Hassan}{equal,to}
\end{mlsysauthorlist}

\mlsysaffiliation{to}{Department of Computer Science and Engieering, Chalmer University and University of Gothenburg, Gothenburg, Sweden}

\mlsyscorrespondingauthor{Muhammad Danish Waseem}{danishm@chalmers.se}
\mlsyscorrespondingauthor{Ahmed Ali-Eldin Hassan}{ahmed.hassan@chalmers.se}


\vskip 0.3in

\begin{abstract}
Vision-language-action (VLA) models map camera observations and language instructions to robot actions.
However, their large backbones and repeated action generation can make inference take longer than the robot real-time requirements for its next action.
To reduce the load of VLA computations, researchers have proposed multiple optimization approaches, including quantization and pruning. These optimizations can be combined with other general optimizations such as better kernel compilation or using CUDA graphs instead of the eager mode. However, most of these optimization results are usually reported on a single GPU type, typically for a few VLA models, with no analysis on what are the real bottlenecks for a model on a particular hardware or how these optimization would generalize beyond the reported results.
In this paper, we provide a comprehensive measurement study of seven families of inference optimization methods, from CUDA graphs and compilation to token pruning and quantization, on seven state-of-the-art VLAs evaluated on four GPU families ranging from a datacenter H100 to an embedded Jetson Thor. We analyse how CPU launch overhead, memory bandwidth and kernel efficiency jointly shape optimization gains, and how applying one method changes the benefit of another.
From this analysis, we derive VLA-Triage, guidelines for stacking optimizations under latency budgets.
The guidelines identify which stages can provide the required saving and which combinations need testing in the deployed execution mode.
\end{abstract}
]



\printAffiliationsAndNotice{}  
\section{Introduction}
\label{sec:intro}
Vision-language-action models (VLAs) are multimodal models that enable robots to translate visual observations and natural-language instructions into executable actions, such as moving an arm or opening a gripper. 
They typically build on pretrained vision-language models that are trained on robot demonstrations, connecting visual and semantic knowledge with the physical skills needed to perform tasks~\cite{zitkovich2023rt}. 

 VLAs aim to support general-purpose robots that can follow diverse instructions and generalize to unfamiliar objects and environments~\cite{openvla,pi05}.

Internally, VLAs typically run a VLM with millions or billions
of parameters to generate the robot's control steps. This limits the reaction time of a robot by the inference latency of the VLA. At each policy inference step, a typical VLA receives a natural-language instruction, represented as text, and image observations from one or more cameras. In a common architecture, a vision encoder converts camera images into visual embeddings, which are projected into the language model’s embedding space and processed jointly with the instruction tokens by a transformer backbone. The resulting representations condition the prediction of a robot action or a short sequence of actions with the action generations performed using an action-head.

VLAs need to be real-time, providing a robot with a sufficient number of actions such that the robot does not stall. Robots typically operate with strictly bounded latencies to maintain a continuous, closed-loop feedback cycle. In addition, if a VLA is too slow, it acts on stale visual data, i.e., by the time the model computes and outputs a command, the physical world has already changed leading to dropped objects, collisions, or unstable movement. This is challenging, especially when the VLA is running on the limited on-robot hardware.
One approach to reduce the computational time of VLAs is to use optimization methods such as, token pruning, quantization, compilation, or CUDA graphs. However, our study of 25 published paper (Appendix \ref{app:audit}) reveals that most of these approaches  do not consider the intricacies of the different VLA architectures and the hardware they run on. 

VLAs today vary in architecture, backbone size, visual token count and action path.
These choices decide which stage dominates latency and therefore which optimization can help.
The backbone size sets how many weight bytes every pass must read, so a large model could spend most of its time reading the model weights, whereas a smaller model could spend its time in the denoising loop of action head. In addition, previous work on VLA optimizations does not consider the inherent bottlenecks of the underlying hardware platform projecting generality of the results while only evaluating the work on a limited set of hardware. For example, many previous works evaluate the optimizations on GPUs such as NVIDIA RTX 4090s, a GPU that has a heavy weight and high power draw---it  is neither designed for running on a robot platform nor is available on most clouds, hence, not supporting a cloud-robotic deployment.

In this work, we provide a comprehensive study of VLA optimization performance. We start by analyzing seven families of inference optimization methods, namely; CUDA graphs, compilation, fused attention, batching, reducing the denoising steps, token pruning and quantization. We evaluate these optimizations on seven VLAs, namely; OpenVLA, OpenVLA-OFT, LLaVA-VLA, CogACT, $\pi_{0.5}$, GR00T, and SmolVLA, with four kinds of action head (token decoding, parallel decoding, diffusion, flow matching). These combinations are then tested on four different GPU families; a datacenter grade
H100, a cloud A10G, a consumer RTX 4090, and an embedded Jetson Thor. 

Our evaluations focus on the latency gains of these optimizations, showing that the optimization results are non-uniform. Gains
differ between GPUs in magnitude, in sign and in rank.

For example, compiling GR00T with CUDA graphs makes it faster on the H100 and slower on the Thor, and storing
CogACT's weights in 8 bits makes it faster on the A10G and slower on the H100. In addition, we find that the choice of baseline to compare an optimization method to can be misleading, e.g.,
FastV~\cite{fastv} and several other token pruning methods (Appendix \ref{app:audit}) select the visual tokens to prune by their attention weights, so it is typically
evaluated and compared to running with unfused attention. However, VLAs in production are often deployed with fused attention kernels such as FlashAttention-2~\cite{flashattention2}, which never write attention weights to memory. Measured against unfused attention, FastV at $50$\% pruning saves $21.4$\% of OpenVLA-OFT's inference time on an H100. Measured against FlashAttention-2, the same run saves $6.9$\%, because the selector reads attention weights and forces the slower kernel (Table~\ref{tab:selector}).

Our analysis also shows that each stage of inference is limited by either the GPU compute power, its memory bandwidth, or the CPU that
launches the GPU kernels. This means that the same stage can be subject to a different bottleneck on different GPUs. Therefore, a proposed optimization method's benefit may not extend beyond the platform it is tested on. 

In addition, optimization methods can be combined, or stacked. Hence, running multiple optimizations in sequence means that later VLA optimizations act on the inference stages that are left by previous optimizations. Hence, the total gains from a later method depends on the methods applied before it. Based on the above observations, we formalize \emph{VLA-Triage}, a systematic approach for evaluating the expected gains of multiple VLA optimization methods on a given hardware. To summarize, our paper has three connected contributions:
\begin{itemize}\setlength\itemsep{2pt}
\item \textbf{Characterisation}: paired measurements of inference-stage costs and optimization benefits across seven VLAs and four GPU-host platforms, including cases where a method helps on one platform and hurts on another.
\item \textbf{Interactions}: We show how graphs, compilation and pruning change the gains for later applied methods, and how reusing standalone gains can change latency-budget decisions.
\item \textbf{Guidelines}: A measurement-based workflow for choosing an execution mode, checking feasibility under stated assumptions, and validating method combinations to be used in deployment.
\end{itemize}

\section{Background}
\label{sec:background}
\label{sec:related}

\label{sec:background}
\label{sec:related}

\begin{table*}[t]
\centering
\small
\caption{The seven evaluated VLAs. Token counts cover all cameras and total prefill. AR: autoregressive; DiT: diffusion transformer; CFG: classifier-free guidance. $A$: actions executed per inference; $A_r$: actions queued when requesting the next chunk; $B$: latency budget (Section~\ref{sec:background}). CogACT executes one action per inference despite predicting a chunk. Kernel counts and $T$ use eager bf16, batch one, on H100.}
\label{tab:models}
\label{tab:budgets}
\begin{adjustbox}{max width=\textwidth}
\begin{tabular}{lllllrrrrrrr}
\toprule
Model & Family & LLM & Vision & Action path & vis.\ tok. & prefill tok. & $A$ & $A_r$ & $B$ (ms) & kernels & $T$ (ms) \\
\midrule
OpenVLA & decoding & Llama-2 7B & DINOv2+SigLIP & AR decode, 7 tok. & 256 & 291 & 1 & -- & 50 & 12k & 103.8 \\
OpenVLA-OFT & prefill-heavy & Llama-2 7B & DINOv2+SigLIP & parallel + MLP & 512 & 605 & 8 & 3 & 150 & 2.7k & 42.4 \\
LLaVA-VLA & decoding & Vicuna 7B & CLIP-L/336 & AR decode, 37 tok. & 576 & 633 & 5 & 2 & 100 & 60k & 483.6 \\
CogACT & prefill-heavy & Llama-2 7B & DINOv2+SigLIP & DiT, 10 DDIM$\times$CFG & 256 & 276 & 1 & -- & 50 & 4.4k & 50.7 \\
$\pi_{0.5}$ & flow & Gemma 2B & SigLIP-So400m & flow expert, 10 steps & 512 & 712 & 10 & 4 & 200 & 13k & 111.6 \\
GR00T N1.7 & flow & Qwen3-VL 2B & Qwen3-VL ViT & DiT flow, 4 steps & 128 & 156 & 16 & 7 & 350 & 7.7k & 74.9 \\
SmolVLA & flow & SmolVLM2 0.5B & SigLIP & flow expert, 10 steps & 128 & 149 & 50 & 24 & 1200 & 26k & 208.3 \\
\bottomrule
\end{tabular}
\end{adjustbox}
\end{table*}

\begin{figure}[t]
\centering
\includegraphics[width=0.84\columnwidth]{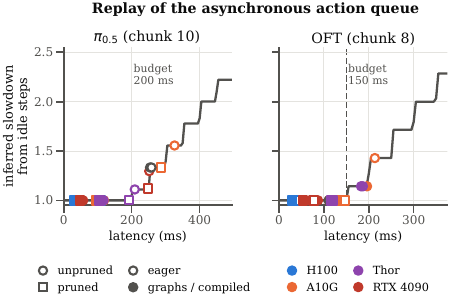}
\caption{Inference latency against the robot's budget.
Replay of asynchronous action queue. Points are measured latencies and dashed lines the budgets $B$ of Equation~\eqref{eq:budget}. Beyond $B$ the queue finishes before the next action chunk arrives; the vertical axis is the resulting slowdown. }
\label{fig:teaser}
\label{fig:flips}
\end{figure}
\subsection{VLA Computation and Time Budgets}
VLA inference has three main stages. First, a vision encoder converts images into visual tokens. After that, the language backbone  processes these tokens together with the task instructions; this stage is called prefill. Finally, an action-generation stage produces the robot's commands.
A chunk is a sequence of actions produced by one inference, so the robot can execute several control steps before requesting another prediction.
One of the main differences between VLA models is in how they generate actions. This can be categorized in three main model categories:

\begin{enumerate}\setlength\itemsep{1pt}
\item \textbf{Autoregressive decoding models}, e.g., \emph{OpenVLA}~\cite{openvla} and \emph{LLaVA-VLA}~\cite{llavavla}: Represent actions as a sequence of tokens, generated one per forward pass. Repeated passes read the language-model weights, making memory transfers, CPU kernel launches, and kernel performance important. 

\item \textbf{Prefill-heavy models}, e.g., \emph{OpenVLA-OFT}~\cite{openvlaoft} and \emph{CogACT}~\cite{cogact}: Run the language backbone once during prefill. OpenVLA-OFT predicts a chunk using action placeholder tokens processed in parallel. CogACT feeds prefill embeddings into a small diffusion transformer sampled with DDIM, a diffusion sampling method~\cite{song2021ddim}. 

\item \textbf{Iterative flow models}, e.g., \emph{$\pi_{0.5}$}~\cite{pi05}, \emph{GR00T}~\cite{bjorck2025gr00t}, \emph{SmolVLA}~\cite{shukor2025smolvla}: Refine an action chunk through multiple flow-matching steps~\cite{lipman2023flow}, using a small action network. We measured with four steps for GR00T and ten for $\pi_{0.5}$ and SmolVLA.
\end{enumerate}

The inference deadline depends on how the robot consumes its actions.
A single-action policy must finish inference within one control period $\Delta$, which is set by the robot motor frequencies.
With asynchronous execution, a chunked policy computes the next chunk while the robot executes actions already in an action queue~\cite{black2025rtc}. This can lead to the VLA acting on stale video data. Hence the number of actions should not be too large such that the robot will fail due to working on stale video data, or too few such that the robot movements are sluggish due to the VLA inference time.

We model the chunking policy through the asynchronous scheduler of LeRobot~\cite{cadene2024lerobot}, which initiates inference when $A_r$ actions remain in the robot's execution queue.
For comparison, in our experiments, we request the next chunk when roughly half the current chunk remains: $A_r=\lfloor(A-1)/2\rfloor$ for chunk length $A$. The minus one accounts for discarding the action aligned with the observation time.
The queued actions provide time for the next inference. With control period $\Delta=$$50$ ms ($20$ Hz), used in all our experiments, this window is
\begin{equation}
B = \begin{cases}
\Delta, & \text{single-action policy},\\
A_r\Delta, & \text{asynchronous chunked policy}.
\end{cases}
\label{eq:budget}
\end{equation}
Budgets range from 50\,ms to 1200\,ms (Table~\ref{tab:models}), based on the number of actions generated simultaneously by the VLA.
We compare measured inference time $T$ with $B$.
Serial preprocessing of duration $t_{\rm preproc}$ leaves $B-t_{\rm preproc}$ for the model.
If inference exceeds this window, the robot exhausts queued actions and spends a fraction $q_{\rm idle}$ of control steps idle.
Figure~\ref{fig:teaser} plots the resulting inferred slowdown $1/(1-q_{\rm idle})$ for executing the same action trajectory.

\begin{figure}[t]
\centering
\includegraphics{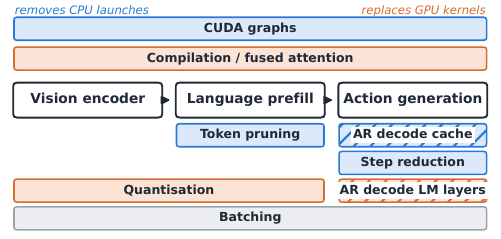}
\caption{Where the optimizations act in VLA inference. Pruning reduces tokens only in LLM Prefill. VLA quantization methods usually quantise the vision encoder and language model and keep a separate action head in full precision, because it is most sensitive to quantization error ~\cite{qvla,actquant}. The hatched band marks autoregressive decoding, which runs the language model.}
\label{fig:pipeline}
\end{figure}

\subsection{Optimizations and Hardware Boundaries}
Figure~\ref{fig:pipeline} maps the methods to the work they change.
For interpreting performance, we distinguish methods that remove repeated work from those that change the GPU kernels used.
\emph{Work-removing methods} eliminate repeated operations.
CUDA graphs~\cite{cudagraphsdoc} reduce repeated CPU launches by replaying recorded kernel sequences.
Visual token pruning~\cite{fastv} removes image tokens after early layers. Step reduction~\cite{tang2026fewersteps,snapflow} removes denoising passes in the action head.
\emph{Kernel-replacing methods} change how the GPU executes the computation.
\texttt{torch.compile}~\cite{ansel2024pytorch2} fuses operations and generates kernels.
Fused attention avoids writing large attention matrices to memory.
Quantization~\cite{qail} compresses weights (int8, int4) or uses lower-precision arithmetic (FP8, int8).
A quantization scheme specifies weight and activation precision, together with a scaling rule.
Weight-only quantization retains bf16 (16-bit floating-point) activations; int8 compute uses int8 weights and activations for matrix multiplication.
FP8 and NVFP4 use 8-bit and 4-bit floating-point values, respectively.
Different quantization schemes can use different kernel libraries.
Batching processes requests together, sharing weight reads to improve throughput.

We call graphs, compilation and fused attention \emph{accuracy-neutral} because they preserve the intended computation up to numerical rounding.
We call pruning, step reduction, and quantization \emph{accuracy-risking}, as they change the computation and require task-specific accuracy evaluation.

\section{Methodology}
\label{sec:eval}
\label{sec:method}

\subsection{What We Measure}
We evaluate the inference optimization methods shown in Figure~\ref{fig:pipeline}, with seven VLA models listed in 
Table~\ref{tab:models}. Table~\ref{tab:platforms} in Appendix lists the four GPU-host systems we tested with, along with their software versions, measured throughput, bandwidth and CPU launch costs.

We measure the time starting after CPU image preprocessing and ending when the model returns its actions.

We validate that the model's input and output remain valid during measurements (details in Appendix A).

\paragraph{Execution modes.}
We compare three execution modes. 1)~In eager execution, PyTorch launches operations as they run. 2)~CUDA graphs replay recorded GPU work with a single CPU launch. 3)~\texttt{torch.compile} is used with its \texttt{reduce-overhead} option, which also uses CUDA graphs where supported, and its \texttt{max-autotune} option, which also benchmarks alternative matrix-multiplication kernels.
The Llama-family VLA stages are compiled separately so that CUDA events can still time them, with decode served by one graph over a fixed-size masked cache. OpenPi ships $\pi_{0.5}$ with the whole call compiled, and we keep that scope. For GR00T we compile the DiT action loop, which takes most of the inference time, and leave the Qwen3-VL backbone eager. For SmolVLA, compiling the whole call has no effect. Its vision embedding selects tokens with a mask computed from the image, so the resulting shape is only known at runtime. The compiler cannot trace such an operation, and \texttt{torch.compile} would then run the entire function eagerly. We therefore compile only the denoising step, which runs ten times per inference.

We keep each VLA's attention backend as released. The Llama models use FlashAttention-2~\cite{flashattention2}, GR00T and SmolVLA use scaled dot-product attention (SDPA), and $\pi_{0.5}$ uses unfused attention with OpenPi's explicit prefix-LM mask, which the fused kernels do not accept.
Accuracy-neutral modes are compared against eager execution.
Token pruning and step reduction are evaluated against the unpruned and full-step baseline, respectively, in the same execution mode.

Quantization is compared in two execution modes. The main comparison (Table~\ref{tab:formats}) uses reduce-overhead compilation, with the bf16 baseline measured in the same mode as the quantised runs. A separate comparison measures the same schemes under CUDA graphs without compilation, to separate the gain of the quantised kernels from the gain the compiler adds by fusing their scaling and casting. Only language-backbone linear layers are quantised.
For each quantization scheme and platform, we time the available library implementations and use the fastest one whose output error stays within 1.25 times the smallest error among them (Table~\ref{tab:survey}).

\paragraph{Timing protocol and error thresholds.}
Short GPU kernels make latency sensitive to CPU launch delays.
We pin the issuing process to an idle CPU core near the GPU and warm up before timing.
Each measurement round repeats the job's selected configurations.
We report medians of at least 200 inferences per configuration, or 100 batches for batching.
Across repeated jobs, the median latency difference is 0.2\% under CUDA graphs and 1.4\% in eager execution (Appendix~\ref{app:repro}).
The largest eager baseline change we observed, when remeasuring the RTX 4090 baselines, is 5.3\% (Appendix~\ref{app:repro}).

\subsection{What Each Optimization Changes}
\label{sec:principle}

We test whether a saving measured before an optimization remains applicable after it. We call this measurement reuse. We ask whether the earlier optimization changes the GPU work.
We call a change \emph{execution-preserving} if it mainly removes CPU launches while retaining GPU kernels and their input dimensions, otherwise it is \emph{execution-altering}.
Compilation, pruning and quantization fall in the latter group.
The distinctions serve different purposes. Accuracy-neutral or risking indicates whether task validation is needed. Work-removing or kernel-replacing describes what the method changes. Execution-preserving or altering indicates whether eager measurements may remain informative.

\paragraph{Accounting for stage gains.}
If an optimization changes only the stage it targets, accelerating a fraction $\sigma$ of baseline latency by a factor $s$ saves an overall fraction $g = \sigma (1 - 1/s)$ of baseline time.
Here the stage share $\sigma$ is the fraction of inference time spent in the affected stage, and $s$ is its speed-up.
We use the identity to explain measured changes.
An earlier optimization can change the stage share $\sigma$, the speed-up $s$, or both. For instance, once compilation shortens the action loop, it lowers action head share $\sigma$, leaving fewer denoising steps less time to save. Similarly, pruning lowers the share of prefill layers and also changes their matrix sizes, and with them the speed-up $s$ of the quantised kernels applied to them. Section~\ref{sec:stacking} examines these effects.

\subsection{Stage Timings and Hardware Reference Costs}
We record stage times using CUDA events.
Separate CUDA Profiling Tools Interface (CUPTI) traces record kernel busy time $K$: the total GPU duration during which at least one kernel is executing.
We estimate whole-inference time outside kernels as $I = T - K$, using unprofiled latency for $T$ to limit profiler host-side instrumentation overhead.

For each matrix or attention operator $o$ in stage $j$ of a VLA, we count its floating-point operations $F_o$ and estimate its memory traffic $W_o$.
With the platform's measured peak compute throughput $C$ and DRAM bandwidth $BW$ (Table~\ref{tab:platforms}), the roofline reference time for stage $j$ is:
\begin{equation}
L_j = \sum_{o \in j} \max\left(\frac{F_o}{C}, \frac{W_o}{BW}\right)
\label{eq:split}
\end{equation}

Measured time can exceed $L_j$ because of uncounted traffic, launch gaps and inefficient kernels.
Optimization decisions rely on direct measurements on the target. Appendix~\ref{app:extra} details the reference calculations.

The roofline reference~\cite{williams2009roofline} (Figure~\ref{fig:roofline}) compares arithmetic intensity (FLOPs per byte) with the ridge point $C/BW$, where the idealised memory and compute references cross.
For a bf16 linear layer with $n$ parameters and $M$ tokens, $2Mn$ operations and $2n$ weight bytes give an intensity near $M$ when weight reads dominate; long prefills can approach the compute limit, while single-token decoding ($M=1$) lies on the memory side .

\section{Bottlenecks of Each VLA Stage}
\label{sec:limits}

\begin{figure*}[t]
\centering
\includegraphics[width=0.8\textwidth]{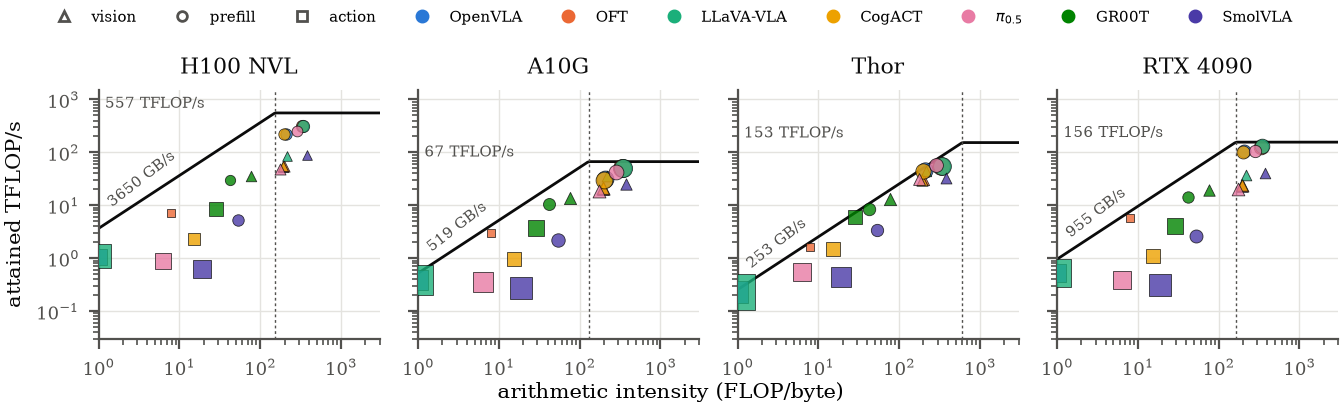}
\caption{Stage throughput against each platform's roofline reference, unpruned bf16 at batch one. Colour denotes the model and shape denotes the stage. Arithmetic intensity uses the operator traffic of Equation~\eqref{eq:split}.}
\label{fig:roofline}
\end{figure*}

Figure~\ref{fig:roofline} shows the roofline model for the different stages of the seven VLAs on the four GPUs. Autoregressive decoding approaches the memory reference on A10G and Thor. H100 leaves more headroom.
On the H100, a decoded Llama-2-7B token takes 9.6\,ms under graphs (Table~\ref{tab:costs}) against a 3.7\,ms weight-read reference at the measured bandwidth. Each pass issues well over a thousand kernels. Compilation with max-autotune reduces kernel counts and brings the time per decoded token to 5.7\,ms. Many of these kernels are short elementwise and normalisation operations that move little data and leave DRAM bandwidth underused, so fusing them closes part of the gap. The measurements do not isolate how much of the remaining gap each kernel accounts for.
All Thor stages lie left of the ridge point under this traffic estimate, where memory cost exceeds compute cost in the reference. 

The gap to measured performance can include kernel inefficiency, CPU launches and differences in actual traffic, so position in Figure~\ref{fig:roofline} alone does not establish the bottleneck.
Appendix Figure~\ref{fig:stagelimits} helps interpret stage costs using measured graph savings, estimated overhead for flow models, compute and memory reference costs, and the unexplained remainder. The flow-model estimates do not isolate CPU waiting. Graphs can reduce launch delays, pruning can reduce token-dependent work, and quantization can reduce weight traffic or change arithmetic costs. Their opportunities therefore depend on which costs remain.

\subsection{CPU Launch Costs and GPU Kernels}
\label{sec:neutral}

\paragraph{Reducing time outside kernels.}
For the Llama-based models, CUDA graphs save more where the estimated idle share $I/T$ is larger (Appendix Figure~\ref{fig:graphs}). The idle share is the fraction of eager inference time in which no GPU kernel is running.
On the A10G and Thor, decode loops spend more time reading weights and gain little from graphs.
The saving depends on whether the CPU can launch the next kernel before the GPU finishes its current work.

\begin{figure*}[t]
\centering
\includegraphics[width=0.8\textwidth]{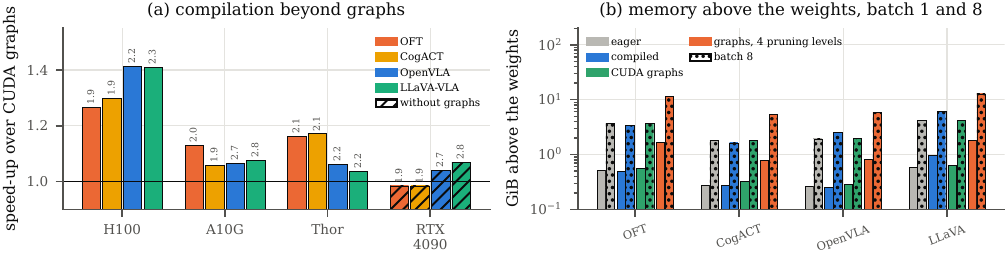}

\caption{Latency benefits and memory costs of accuracy-neutral methods.
\textbf{(a)} Speed-up of reduce-overhead compilation over CUDA graphs alone for the four Llama-based models. The RTX 4090 shows compilation without graphs. Bar labels give the eager kernel count divided by the compiled count.
\textbf{(b)} GPU memory beyond the model weights at batch sizes one and eight, medians over the GPUs where the configuration fits. The CUDA-graphs bars hold one graph. The four-level bars hold one graph per pruning level, each with its own memory pool, and at batch eight OpenVLA-OFT and LLaVA-VLA then exceed the 24GB of an RTX 4090 or A10G.}
\label{fig:neutralapp}
\end{figure*}

\paragraph{Compilation and fusion.}
Compilation fuses elementwise and normalisation operations, reducing kernel counts by $1.9\times$ to $2.8\times$ in our experiments, and providing additional speed-ups after CUDA graphs reduce time outside kernels. Figure~\ref{fig:neutralapp}a shows both effects. Its bar heights give the speed-up over graphs alone, and its bar labels the ratio of eager to compiled kernel counts. Fewer kernels also means fewer launches, so the speed-up may come from launch and scheduling savings as much as from reduced DRAM traffic.
Compilation also changes the preferred quantization scheme.
For OpenVLA-OFT on H100, static FP8 is fastest under compilation, speeding up whole inference by $1.43\times$ over compiled bf16 and prefill alone by $1.68\times$ (Table~\ref{tab:formats}).
Under graphs without compilation it is instead the slowest of the three, at 33.5\,ms against 31.5\,ms for dynamic FP8 and 31.0\,ms for int8 compute.
Compilation changes two things at once. It fuses the scaling and casting of activations into neighbouring kernels, and it can select different matrix-multiplication kernels.

\paragraph{Compilation slowdowns and resource costs.}
Compilation can also slow inference. For GR00T on Thor, compiler-generated copy and conversion kernels increase GPU kernel duration despite shorter gaps between kernels (``GR00T compiled on the Thor'' in Appendix~\ref{app:groot}).
Execution modes also require memory and setup time.
Each captured graph uses fixed tensor dimensions and reserved GPU memory pools, while compilation can take minutes (Appendix Table~\ref{tab:setup}).
Figure~\ref{fig:neutralapp}b shows the memory each execution mode needs beyond the model weights. A CUDA graph is captured for fixed tensor shapes, so a deployment that varies its batch size or pruning level at run time must keep one graph per shape resident. Each graph reserves a memory pool roughly as large as the activation memory of eager inference at that batch size. At batch eight, four pruning-level graphs held at once reserve 11 to 13\,GB of pools on top of 13 to 14\,GB of weights for OpenVLA-OFT or LLaVA-VLA, which exceeds the 24\,GB of an RTX 4090 or A10G.

\subsection{Tokens, Steps and Quantization}
\label{sec:risking}

\paragraph{Token pruning.}
Pruning reduces token-dependent work in the language-model layers. We use random token selection to measure the latency effect of token count.

These timings measure performance at each token count. They do not determine which token-selection method preserves task accuracy.
Under CUDA graphs, prefill latency generally shortens with token count (Appendix Figure~\ref{fig:prefill}).
Pruning reduces token-dependent computation but does not eliminate the weight reads of layers that still run.

\emph{Fewer tokens do not always reduce latency.}
Matrix-multiplication (GEMM) kernels process tokens in fixed-size blocks called tiles, so removing tokens that do not empty a whole tile saves little.
Some measured points even become slower as the token count falls (Appendix Figure~\ref{fig:prefill}), likely because the library selects a different kernel at the new shape. 
In eager execution, pruning can also stop helping once CPU launches, rather than GPU work, set the latency.

\emph{Token selection adds its own cost.}
FastV~\cite{fastv} ranks visual tokens by attention weights.
Obtaining those weights requires a slower, unfused attention implementation. 

Table~\ref{tab:selector} compares these variants with random selection, showing that selection overhead and the attention backend directly affect the realised latency saving.

\paragraph{Fewer denoising steps.}\label{sec:steps}
In flow and diffusion heads, each denoising step repeats a forward pass over identical action-head weights.
Fewer steps shorten this loop while leaving vision and prefill unchanged.
Appendix Table~\ref{tab:costs} separates the cost of one denoising step from the time outside the denoising loop.
In eager execution, small action heads issue many short kernels, leaving the GPU waiting for CPU launches.
Compilation accelerates individual steps in $\pi_{0.5}$ and SmolVLA by over $2\times$, whereas GR00T's generated kernels on Thor lengthen step time.

\paragraph{Quantization.}\label{sec:quant}
Quantization stores weights at lower precision. Quantization schemes that also quantize activations use lower-precision inputs for matrix multiplication.
As shown in Table~\ref{tab:formats}, weight-only quantization schemes (int8, int4) accelerate memory-bandwidth-bound stages, cutting decode loop latency on the A10G and Thor.
During prefill, dequantization and kernel overhead can outweigh the time saved by reading fewer weight bytes.
Activation quantization schemes (FP8, int8 compute) use lower-precision tensor cores to accelerate dense prefill GEMMs.
They need enough tokens for this saving to outweigh quantization and scaling overhead.
A weight-only quantization scheme can therefore shorten decode while lengthening prefill in the same inference.

\section{How One optimization Changes the Next}
\label{sec:stacking}

\label{sec:stacking}

\begin{figure*}[t]
\centering
\includegraphics[width=0.70\textwidth]{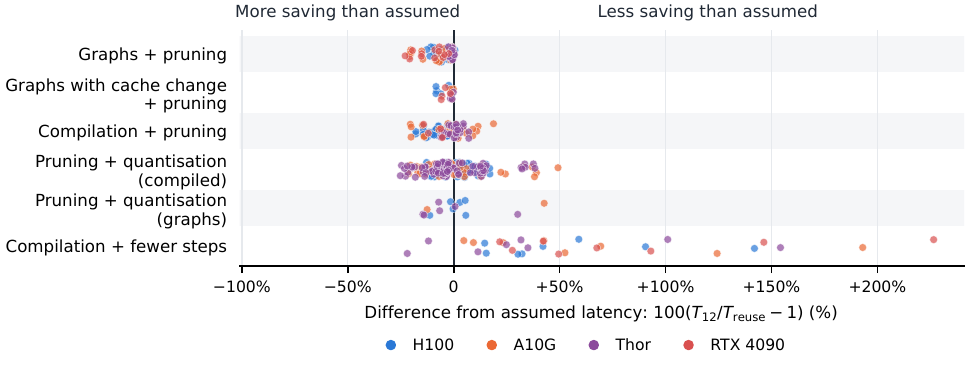}

\caption{Testing whether a separately measured percentage saving carries over to a combination. Each point compares one measured combination $T_{12}$ with the latency implied by the separate measurements, $T_1T_2/T_0$, for one model, platform and setting, such as a pruning level, quantization scheme or step count. Zero means the percentage saving is unchanged. Positive values mean less saving than assumed, and negative values mean more. Appendix Tables~\ref{tab:census_reuse}--\ref{tab:census_flips} give the counts and millisecond-reuse results.}

\label{fig:reuseoverview}
\end{figure*}

A deployment may combine methods whose benefits were measured separately. We ask whether a method's saving still holds once another optimization has been applied.
We use the distinction between execution-preserving and execution-altering from Section~\ref{sec:principle} to interpret the results.

\paragraph{Measuring the change in saving.}
On a fixed model, platform and software version, we measure the baseline $T_0$, the earlier optimization alone $T_1$, the later method alone $T_2$, and both together $T_{12}$.
The later method saves $T_0-T_2$ without the earlier optimization and $T_1-T_{12}$ with it.
We test two assumptions about whether that saving remains unchanged.
\emph{Percentage reuse} assumes the later method saves the same percentage of time, giving $T_{\rm reuse}=T_1(T_2/T_0)$.
\emph{Millisecond reuse} assumes it saves the same number of milliseconds, giving $T_{\rm reuse}=T_1-(T_0-T_2)$.
We compare the latency implied by each assumption with the measured $T_{12}$. An assumption is \emph{optimistic} when the implied latency is lower than measured, and \emph{pessimistic} when it is higher.
We report cases where measured latency exceeds the implied value by more than 5.3\%, the largest eager baseline change observed during remeasurement (Section~\ref{sec:method}).

\subsection{Three Interaction Cases}
\label{sec:interactioncases}
We perfrom a case-study on three example and likely optimization sequences (summarized in Figure~\ref{fig:reuseoverview}). These cases examine the selected sequences, not all possible optimization combinations.

\paragraph{Case 1: CUDA graphs, then pruning.}
CUDA graphs leave the GPU kernels unchanged. Pruning therefore removes the same token-dependent work as in eager execution. Graphs also remove launch overhead, so total inference is shorter. The kernel time that pruning removes is then a larger share of that shorter total. In eager execution part of that kernel time was hidden behind CPU launches, so the milliseconds saved change as well. The eager percentage therefore does not carry over exactly.
In the recorded comparisons, measured latency under CUDA graphs ranges from $23$\% below to less than $1$\% above the value implied by applying pruning's eager percentage saving unchanged.
Profiles show that the small excesses involve changes in both kernel time and time outside kernels. Appendix~\ref{app:closure} splits each excess into a kernel-time term and a time-outside-kernels term from the traces, and shows that both stay within about 1\,ms on the H100 and A10G and largely cancel on the Thor.

LLaVA-VLA needs separate treatment because CUDA graphs change its decoding cache. Its decoding attends over a fixed-size masked cache, changing kernel timing (Section~\ref{sec:principle}). LLaVA-VLA's pruning savings must therefore be measured under CUDA graphs rather than carried over from eager execution.
Figure~\ref{fig:reuseoverview} plots LLaVA-VLA in the ``Graphs with cache change'' row.
OpenVLA uses the same cache layout, but its recorded timing changes are small, so we treat it as approximately execution-preserving.

\paragraph{Case 2: Compilation, then pruning or fewer denoising steps.}
Compilation changes kernel execution times. This changes how much time pruning or fewer steps can save.
For CogACT, 75\% pruning saves 3.9\,ms on H100 in eager execution and 4.1\,ms after max-autotune compilation. Prefill is a small share of CogACT's time on H100 in both modes, so pruning has little to remove either way.
However, on A10G, CogACT's saving falls from 74.2 to 33.5\,ms as compilation already shortens prefill (Appendix Table~\ref{tab:hardwareinteraction}).
Compiler mode also matters: switching unpruned CogACT from reduce-overhead to max-autotune on A10G saves 42.7\,ms, more than the subsequent pruning saves.

Of 84 compilation-and-pruning configurations, eight take over 5.3\% longer than percentage reuse implies: six on A10G and two on Thor. The largest discrepancy is 19\%.
The assumption can also understate savings, as in every H100 comparison in Figure~\ref{fig:reuseoverview}.

For $\pi_{0.5}$ and SmolVLA, compilation accelerates denoising more than the remaining inference. Removing steps then saves less time. GR00T on Thor shows the reverse: compilation lengthens its steps (Section~\ref{sec:steps}).

In 29 of 32 flow-model comparisons, latency is more than 5.3\% higher than the percentage reuse implies, reaching up to 227\% higher.
For SmolVLA on RTX 4090, the assumption gives 17.4\,ms versus 56.9\,ms measured.
Carrying over the eager millisecond saving gives zero or negative latency in 21 comparisons.
Pruning and step-reduction savings therefore need measurement in the chosen compiler mode.

\paragraph{Case 3: Pruning, then quantization, with execution mode fixed.}
Pruning is work-removing because it reduces token-dependent computation. The smaller matrices can nevertheless change kernel selection and efficiency.
Quantization's benefit can change in two ways. Pruning can reduce the share of inference spent in the layers quantization accelerates, while the changed matrix sizes can raise or lower the speed-up of quantised kernels.
For example, on A10G with reduce-overhead compilation, 75\% pruning reduces the prefill speed-up from int8 compute for CogACT but increases it for OpenVLA-OFT (Appendix Figure~\ref{fig:quanttokens}, Table~\ref{tab:formatinteractionfull}).

In the 194 compiled pruning--quantization configurations, measured latency exceeds the value implied by unchanged quantization percentage savings by more than 5.3\% in 52, including 24 for OpenVLA-OFT on Thor. The largest discrepancy is 49\%. Other configurations have lower measured latency than the assumption implies.
In comparisons retaining individual token selectors, updating the affected-stage share while assuming the unpruned quantization speed-up still leaves measured latency up to about 39\% above the implied value.
Both the affected-stage share and quantization speed-up therefore matter (Appendix~\ref{app:census}).
quantization should be evaluated at the retained token count and in the deployed execution mode.

\subsection{Consequences for Latency Budgets}

A change in saving can determine whether a combination meets its inference-time budget.
For CogACT with 75\% pruning after max-autotune compilation on A10G, assuming unchanged eager millisecond savings gives 23.3\,ms, below the 50\,ms budget. The measured combination takes 64.1\,ms and misses it.
Assuming unchanged percentage savings gives 61.4\,ms, also above the budget.
On H100, compilation alone takes 21.5\,ms and meets the budget without pruning.

Pruning and quantization can change budget decisions too.
For CogACT on A10G with int4 weights and 75\% pruning in reduce-overhead mode, assuming unchanged quantization percentage savings gives 48.5\,ms, within the 50\,ms budget. The combination measures 59.3\,ms and misses it.
With 87.5\% pruning in max-autotune mode, the same assumption gives 59.2\,ms, above the budget, while the combination meets it at 49.2\,ms.
Using the same assumption would therefore lead to the wrong budget decision in both examples.
Appendix Table~\ref{tab:interactionabsolute} gives further examples from earlier measurements, and Appendix Table~\ref{tab:census_flips} gives the full budget comparison.
These checks concern inference time. Section~\ref{sec:procedure} adds reserves for preprocessing, communication, and scheduling.

\subsection{Across Platforms}

We also examine whether a method's measured speed-up persists across platforms.
Among the selected model--method settings, none of the 14 work-removing comparisons changes from a speed-up to a slowdown across platforms, while 6 of the 21 kernel-replacing comparisons do.
With matching PyTorch versions on H100 and A10G, the counts are 0 of 14 work-removing comparisons and 3 of 13 kernel-replacing comparisons. Host CPUs and other software components still differ, so these comparisons do not isolate GPU architecture.
At 25\% pruning, compiled $\pi_{0.5}$ becomes faster on the other tested platforms but slower on Thor. The Thor slowdown appears in two separate measurement jobs.

These cases show that a later method's benefit depends on how much of inference it can shorten and how efficiently the remaining work executes.

\section{VLA-Triage: Guidelines for Stacking optimizations}
\label{sec:triage}
\label{sec:procedure}
\label{sec:triage}

\begin{figure}[t]
\centering
\includegraphics[width=\linewidth]{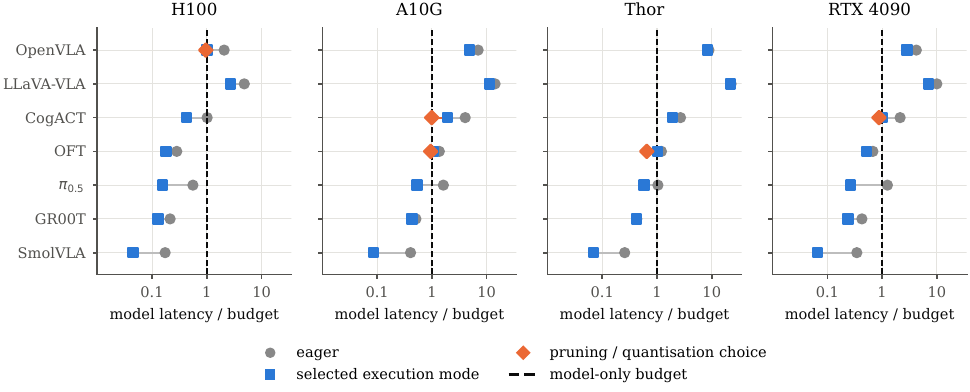}
\caption{Measured model latency relative to the nominal budget, on a logarithmic axis.
Circles show eager execution and squares the selected execution mode. Orange diamonds show one further pruning or quantization setting from Appendix Table~\ref{tab:triage}, choosing pruning when both would qualify.
Points are alternatives, not successive steps. Orange points are omitted when none is needed or none qualifies.
All comparisons exclude deployment reserves.}
\label{fig:ladder}
\end{figure}

VLA-Triage is a set of guidelines based on the findings of Sections~\ref{sec:limits} and~\ref{sec:stacking} that can be used by practitioners for deploying (or evaluating) their VLA optimizations.

\textbf{1.~Set the budget.}
For deployment, reserve time for preprocessing, communication and scheduling delays within $B$:
\begin{equation}
B_{\rm model} = B - t_{\rm preproc} - t_{\rm comm} - t_{\rm margin}.
\label{eq:netbudget}
\end{equation}
Here $B_{\rm model}$ is the time left for model execution, and $t_{\rm margin}$ reserves time for scheduling delays and latency variation. The examples below set $t_{\rm comm}=0$ and $t_{\rm margin}=0.05B$.
Tables that compare measured inference $T$ with $B$ rather than $B_{\rm model}$ exclude these reserves.

\textbf{2.~Choose execution mode.}
Benchmark the supported execution modes (eager, CUDA graphs, and compilation) and attention backends on the physical target.
The attention backend belongs here because it is accuracy-neutral. In the recorded comparisons the best fused backend was never slower than unfused attention on any platform, whether the gain was marginal or substantial, and it lowers attention-heavy prefill, so step 2 selects a fused backend together with the execution mode.
Select the fastest mode that satisfies platform memory and setup constraints (Section~\ref{sec:neutral}).
Use its total latency $T$ and stage timings for the checks in step 3, because eager stage shares need not remain valid after compilation.
If the selected mode alone meets $B_{\rm model}$, skip steps 3 and 4 and go to step 5.

\textbf{3.~Screen unviable candidates.}
\begin{itemize}\setlength\itemsep{1pt}
\item \emph{Weight-read screen.} If reading mandatory weights alone takes longer than $B_{\rm model}$, no faster kernel can make the candidate meet the budget under the same memory traffic.
Let $W_p^{4}$ be the bytes of the vision-language-model weights read in forward pass $p$, counted at four bits, the lowest weight precision among the evaluated  quantization schemes. This gives quantization an optimistic reference, since higher-precision weights require at least as many bytes.
These weights must be streamed from DRAM on every pass. Dividing their volume by datasheet memory bandwidth $BW_{\rm data}$ gives a lower bound on latency:
\begin{equation}
L_{\rm w} = \frac{\sum_p W_p^{4}}{BW_{\rm data}}.
\label{eq:weightread}
\end{equation}
If $L_{\rm w}>B_{\rm model}$, no pruning or quantization setting can meet the budget, since all of them still read these weights. Only candidates that remove reads, by caching weights or dropping layers or passes, remain, and they need their own bound.
\item \emph{Affected-stage screen.} Each method can only shorten the stages it acts on. Pruning acts on prefill. Quantization acts on the vision-language-model layers. Step reduction acts on the action loop of the flow models.
Let $L_m$ be the latency
of method $m$ at its strongest setting. For pruning this is inference with all visual tokens removed, which still runs the vision encoder, the text tokens and decoding. For step reduction it is inference with one denoising step, which still runs vision and prefill. No milder setting is faster, so $L_m$ is a lower bound for method $m$ alone. If $L_m > B_{\rm model}$, skip $m$ as a standalone candidate. A pair of methods acts on more stages than either method alone, so this screen does not exclude pairs.
\end{itemize}

\textbf{4.~Test supported combinations.}

A pruning \emph{level} is the percentage of visual tokens removed. Mild levels are tested first because they risk less accuracy.
If the mode selected in step 2 is CUDA graphs, which largely preserve GPU work, eager pruning timings estimate which levels can meet the budget, through the percentage-saving assumption of Case 1 in Section~\ref{sec:interactioncases}. Select \emph{initial levels} whose implied latency, $T_1T_2/T_0$, is at most $(1+\delta)B_{\rm model}$.
In the recorded comparisons this estimate was usually above the measured latency under CUDA graphs, so a level whose estimate slightly exceeds the budget may still meet it. The allowance $\delta$ keeps such levels in the initial set. We use $\delta=5.3\%$, the largest eager baseline change observed during remeasurement (Section~\ref{sec:method}), and Appendix~\ref{app:screeningobjective} tests other values.

If the selected mode is compilation, eager timings cannot be used to skip evaluations, because Case 2 shows that their savings need not carry over. Measure pruning and step-reduction settings in the compiled mode, from the smallest change upwards.

Quantization schemes are chosen based on hardware support and task accuracy requirements. Measure them in the deployed mode at the retained token count, because pruning can increase or decrease their benefit (Case 3).
If neither pruning nor quantization meets the budget alone, test pairs.

\textbf{5.~Validate task performance.}
For each candidate meeting the latency budget, check success on the chosen benchmark and control stability in deployment.
If either fails, return to step 4 to test further candidates, including further levels or pairs.

\textbf{Meeting the budget with headroom.}
For OpenVLA-OFT on the Thor, the budget is 150\,ms. Reserving recorded serial preprocessing (7.1\,ms) and 5\% of $B$ leaves 135.4\,ms for model execution.
Step 2 selects compilation, with a model latency of 152.6\,ms (Figure~\ref{fig:ladder}).
Removing 50\% of visual tokens drops model time to 97.6\,ms.
Pruning alone therefore meets the latency budget, so the next step is task validation rather than adding quantization.

\textbf{A combination that fails once reserves are included.}
For CogACT on A10G, the 50\,ms budget leaves only 40.5\,ms for inference after 7.0\,ms of serial preprocessing and a 5\% reserve. Max-autotune takes 97.5\,ms. Prefill accounts for 72.3\% of that time, enough that the affected-stage screen does not rule out pruning alone.
Removing 75\% of visual tokens takes 64.1\,ms, and int8 compute alone takes 67.0\,ms. Together they take 49.6\,ms. Although the pair meets the inference-only 50\,ms comparison, it exceeds the time available in deployment.
The fastest recorded setting, 87.5\% pruning with int8 compute, still takes 46.4\,ms. None of the recorded settings meets this deployment budget, so using one would require a longer control period. Appendix Table~\ref{tab:triage} reports the other cases with and without reserves.

Spare latency can support batching or a higher control rate, subject to renewed latency and control checks (Appendix Table~\ref{tab:fleet}).

\section{Related Work}
Latency predictors and tensor compilers already account for operator fusion and shape-dependent kernel selection~\cite{nnmeter,wang2020hat,chen2018tvm,zheng2020ansor}, so a saving measured in one execution mode is known not to transfer to another. VLA acceleration papers usually do not report the mode.  We audit 25 VLA acceleration papers (Appendix~\ref{app:audit}). Only one states the execution mode of both baseline and the method. None of the token-pruning, reuse, merging, or sparsification entries do, and only one of them identifies its attention backend. Readers therefore cannot check whether a reported saving applies to their execution mode.

VLA-Perf~\cite{vlaperf} models architectural and deployment choices with an analytical roofline to estimate ideal compute and memory bounds.
The XPU study~\cite{vlaxpu} benchmarks VLAs across GPUs, an Intel XPU and Ascend NPUs, measuring compilation and a few method pairs, without testing methods against latency budgets.
\citet{jung2026beyondllm} characterise stage balance on an RTX 4090 and two Jetson SoCs and compare per-stage compilation with eager execution, but compilation is the only model-level optimization they measure.

Our paired measurements track how a method's benefit varies across models and GPU-host systems, and how an earlier optimization changes a later method's saving.
We examine graphs before pruning, compilation before pruning or step reduction, and pruning before quantization, then test how the differences affect latency-budget decisions.

\section{Conclusion}

A VLA optimization's benefit depends on the model, GPU-host platform and methods already applied.
Across seven VLAs and four platforms, our measurements show how CPU launch costs, memory costs and kernel efficiency change the opportunity for optimization. Compilation changes pruning and step-reduction savings, while pruning changes quantization benefits.
VLA-Triage formalizes these findings into guidelines for selecting an execution mode, checking feasibility, and validating combinations under a deployment latency budget.
The resulting guidance identifies when standalone measurements remain informative and when the combined execution must be measured.



\bibliography{example_paper}
\bibliographystyle{mlsys2025}

\appendix
\section{Runtime Validation}
\label{app:validation}


\paragraph{CogACT's sampler for CUDA graphs.}
The released DDIM sampler constructs its step schedule on the CPU during each step, preventing this loop from being recorded as a CUDA graph.
Our replacement precomputes the same schedule.
Across five seeds, the maximum absolute difference between its actions and those of the released loop is $1.2\times10^{-6}$ in fp32 on CPU, using identical weights, noise and conditioning inputs.

\paragraph{Token layouts.}
We check that the expected input sequence for each model: 291 tokens for OpenVLA, 605 for OpenVLA-OFT (including 56 action slots), 276 for CogACT (including the cognition token), and 633 for LLaVA-VLA.
We also check LLaVA-VLA's token IDs against its released server using the same instruction.

\paragraph{Scheduler replay.}
We simulate LeRobot's asynchronous action queue, which the robot drains at the control rate while the policy refills it with each inference, to compute how often a given inference time leaves the robot with no action to execute.

\section{Runtimes, Software and Measurement Checks}
\label{app:runtimes}
\label{app:repro}

Timing begins with preprocessed camera frames and ends with actions, excluding CPU image preprocessing.
CUDA events placed between modules divide inference into stages: host-to-device copy, vision encoder, projector, token selection where pruning is enabled, language-model prefill, then decoding for OpenVLA and LLaVA-VLA, the action head for OpenVLA-OFT and CogACT, or the denoising loop for the flow models, and postprocessing. Each stage time is the elapsed time between its two events on the GPU stream.
Compilation uses \texttt{torch.compile}~\cite{ansel2024pytorch2}, and quantisation uses torchao~\cite{torchao}.
Only the language-model linear layers of the Llama-based models are quantised.
For each quantisation scheme and platform, we time the available library implementations and select the fastest one whose output error is within 1.25 times the smallest error measured for that quantisation scheme (Table~\ref{tab:survey}).

On x86 platforms, the Llama-based models and $\pi_{0.5}$ generally use PyTorch 2.7.1.
The exceptions are quantised runs and compilation variants of the Llama-based models on the H100 and A10G, which use PyTorch 2.10 with torchao 0.13.
The Thor uses this stack throughout.
GR00T and SmolVLA use PyTorch 2.11 on x86 platforms and 2.10 on the Thor (Table~\ref{tab:platforms}).
We do not report quantisation schemes on the RTX 4090. 

\paragraph{Timing repeatability.}
When processing one request at a time, latency can depend on which CPU core launches GPU work.
The protocol therefore pins the issuing thread to one core (Section~\ref{sec:method}).
With pinning, OpenVLA's eager rounds agree within 2.1\%.
Across identical configurations measured in separate jobs days apart, the median relative latency difference is 0.2\% under CUDA graphs and 1.4\% in eager execution. The 5.3\% reporting cutoff is the largest eager baseline change observed when remeasuring the RTX 4090 at the end of its campaign. CUDA-graph latencies measured under the two software stacks in Table~\ref{tab:platforms} agree within 4\,ms.
Recalculating each interaction example in Section~\ref{sec:stacking} using every pairing of recorded baseline and optimized round medians preserves the reported direction. OpenVLA-OFT's increase in pruning saving under CUDA graphs on the Thor remains positive in every pairing.

\paragraph{Profiler cross-check.}
We compare kernel busy time from our CUPTI profiler with Nsight Systems, expressing differences relative to the CUPTI value. The differences are at most 0.12\% on A10G and 2.1\% under compilation and CUDA graphs on H100. For eager H100 runs, Nsight reports 4.7--12.2\% more busy time and longer durations for short kernels. On Thor, the differences range from 0.8\% to 2.1\%, and both profilers give the same ordering for all 12 checked configurations.



\section{Additional Measurements}
\label{app:extra}

\begin{figure*}[t]
\centering
\includegraphics[width=\textwidth]{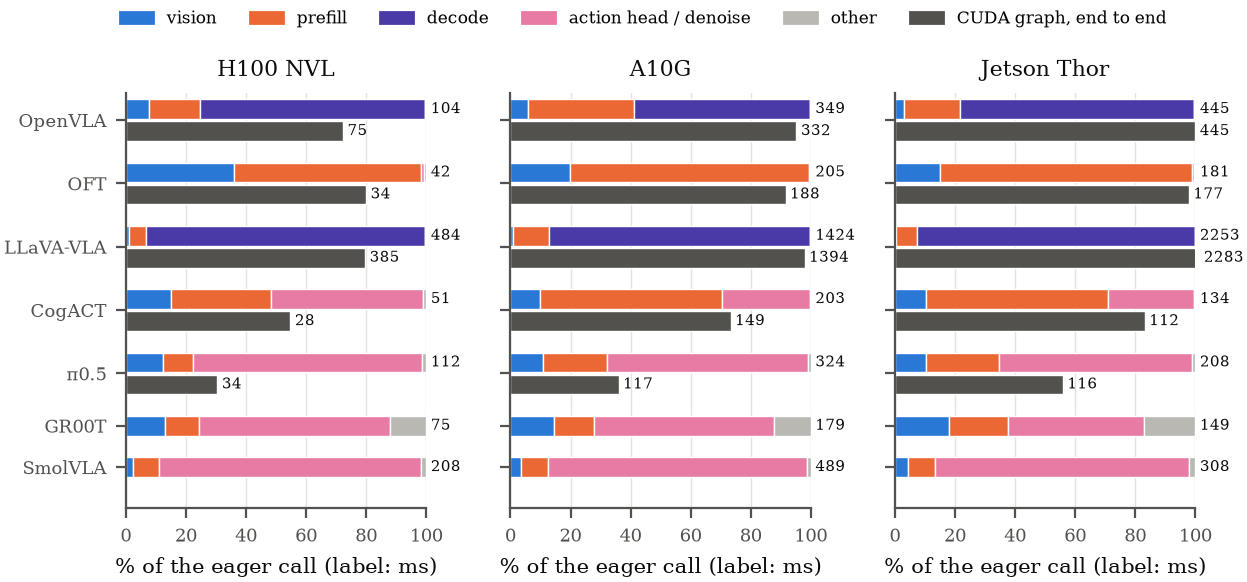}
\caption{Eager inference by stage across models and platforms. Stacked bars show stage shares; labels give whole-inference latency in milliseconds. Dark bars show latency under CUDA graphs, including compilation for $\pi_{0.5}$, relative to the eager total.}
\label{fig:breakdown}
\end{figure*}

\paragraph{Measured stage shares.}
Figure~\ref{fig:breakdown} shows that decoding dominates OpenVLA and LLaVA-VLA. CogACT shifts from a larger action-head share on H100 to a larger prefill share on A10G, changing how much time pruning can remove. Vision includes projection; other includes input transfer, postprocessing and unassigned time.

\begin{figure*}[t]
\centering
\includegraphics[width=\textwidth]{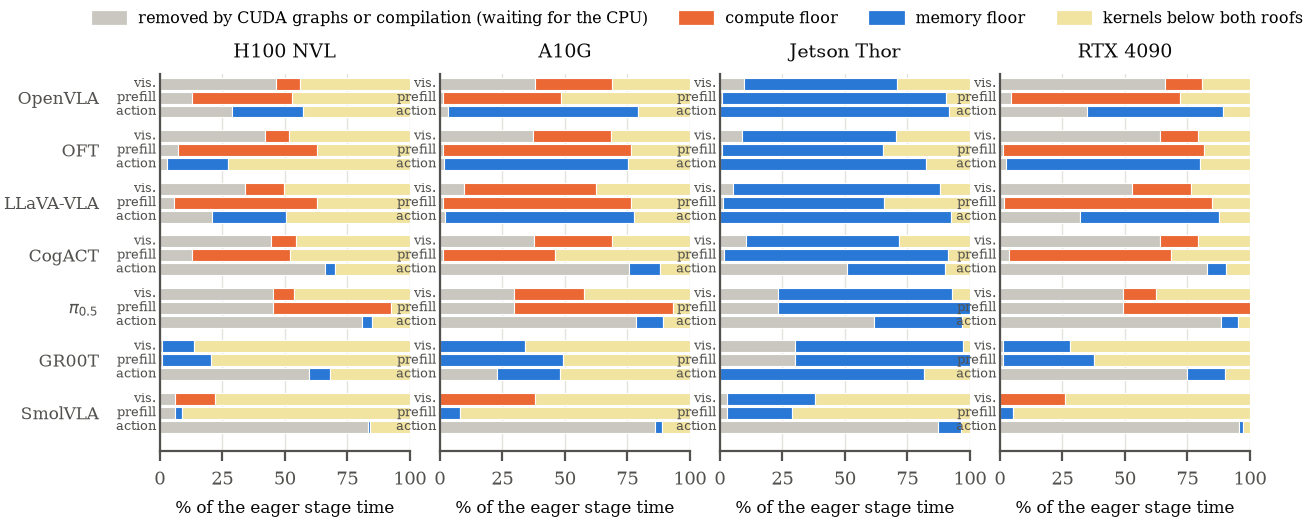}
\caption{Stage-time interpretation across seven VLAs and four platforms. Gray shows measured CUDA graph savings; hatched gray shows estimated removable time for flow models. Orange and blue are compute and memory reference costs; yellow is unexplained time. Allocation and clipping are described in the text. These portions do not establish measured bottlenecks.}
\label{fig:stagelimits}
\end{figure*}

\begin{figure*}[t]
\centering
\includegraphics[width=\textwidth]{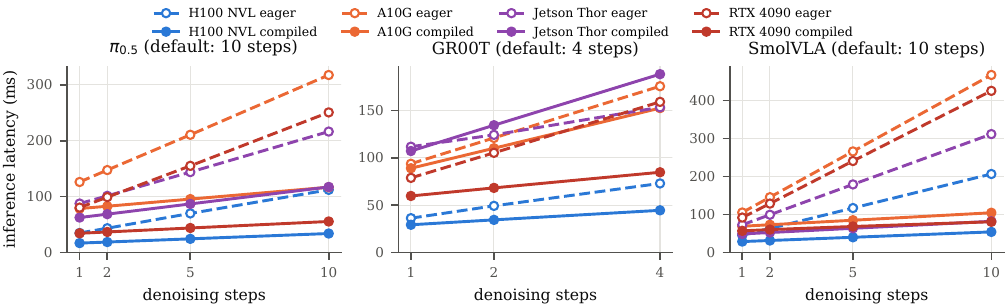}
\caption{Inference latency versus denoising-step count. Colors identify platforms. Dashed lines and open markers show eager execution; solid lines and filled markers show compilation with CUDA graphs. Lines connect measurements. Panel titles give default step counts.}
\label{fig:steps}
\end{figure*}

\paragraph{Reference costs and time outside kernels.}
Figure~\ref{fig:stagelimits} interprets stage time using roofline reference costs and estimated removable time. For the Llama-based models, removable time is the measured CUDA graph saving. For flow models, linear fits over denoising-step counts estimate time per step and time outside the loop (Figure~\ref{fig:steps}). Eager-to-compiled differences supply the removable-time estimate, allocated between vision and prefill in proportion to their eager times. For GR00T and SmolVLA, the removable-time estimate also includes estimated time outside kernels. These allocations do not isolate CPU waiting.

\begin{figure}[t]
\centering
\includegraphics[width=\columnwidth]{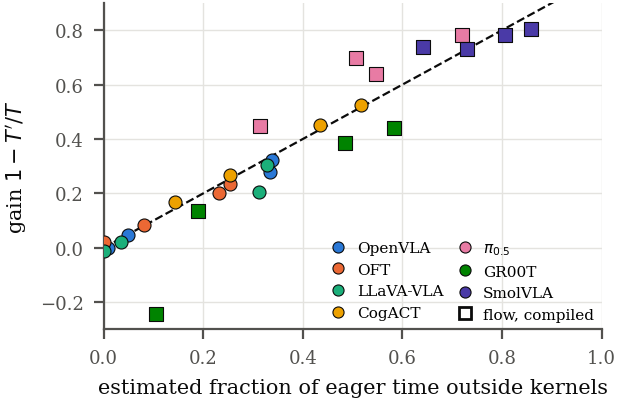}
\caption{Execution-mode savings versus estimated eager time outside kernels, $I/T$. Circles use CUDA graphs alone; squares combine graphs with compilation. The dashed line marks equality. Time outside kernels combines unprofiled latency with a separate CUPTI trace.}
\label{fig:graphs}
\end{figure}

In 6 of 84 bars, the estimated removable time is reduced so that it and the reference fit within measured stage time. The reductions range from 0.29 to 5.02\,ms, or 1.6--23.4\% of stage time. No unscaled roofline reference exceeds measured stage time. Figure~\ref{fig:graphs} separately compares execution-mode savings with the estimated fraction of eager time outside kernels.

\paragraph{Weight-read screen.}
Table~\ref{tab:triage} and the budget-sensitivity check in Appendix~\ref{app:screeningobjective} compare $L_{\rm w}$ of Equation~\eqref{eq:weightread} with the nominal budget $B$, whereas deployment screening uses $B_{\rm model}$ after reserves.
The screen assumes that each counted weight is read from DRAM once per pass. Datasheet bandwidth exceeds the measured bandwidth on every platform, giving a lower transfer-time reference. Exclusion remains conditional on these mandatory reads. Counting the vision encoder at 4 bits also covers methods that quantize it, although our measurements quantize only language-model layers.

\begin{figure}[t]
\centering
\includegraphics[width=\columnwidth]{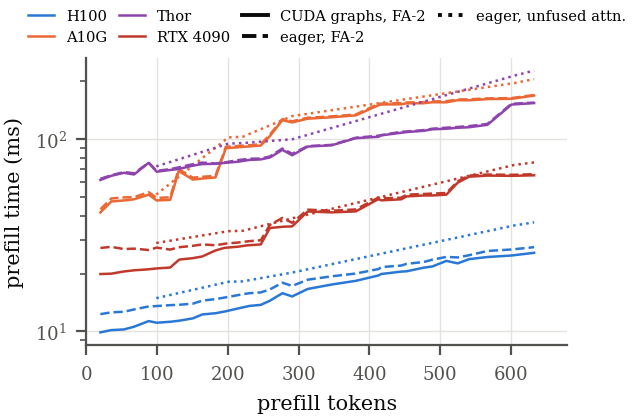}
\caption{Prefill latency versus tokens remaining after layer-3 pruning in the shared 7B Llama architecture. Solid: CUDA graphs with FlashAttention-2; dashed: eager FlashAttention-2; dotted: eager unfused attention. The latency axis is logarithmic. Lines connect measurements.}
\label{fig:prefill}
\end{figure}

\begin{figure}[t]
\centering
\includegraphics[width=\columnwidth]{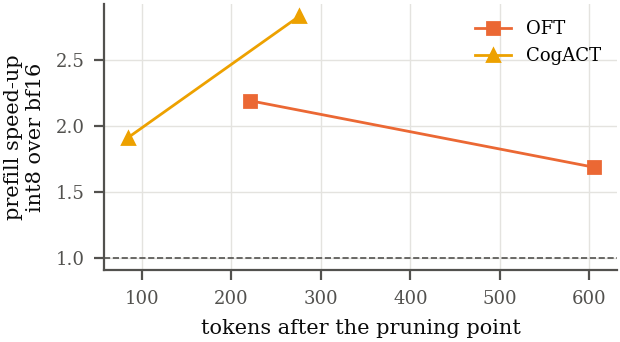}
\caption{Int8 prefill speed-up on A10G at matching pruning levels under reduce-overhead compilation. Earlier layers retain the original token count. Lines connect bf16/int8 comparisons.}
\label{fig:quanttokens}
\end{figure}

\paragraph{Token count and optimization interactions.}
Figures~\ref{fig:prefill} and~\ref{fig:quanttokens} show how token count changes prefill latency and quantization speed-up. Points sharing the 7B Llama architecture contribute to the same platform curve in Figure~\ref{fig:prefill}. Some points slow down as tokens are removed; the plot does not isolate the responsible kernels.
For the int8 comparison, bf16 and int8 use the same pruning level and reduce-overhead compilation. The first three layers retain the original token count, so the post-pruning count does not describe the complete prefill workload.

Tables~\ref{tab:formatinteractionfull} and~\ref{tab:interactionabsolute} retain earlier measurement jobs; the later reuse census (Appendix~\ref{app:census}) repeats measurements and averages paired token selectors. For CogACT on A10G, the reduce-overhead baseline is 140.3\,ms in the earlier table and 140.2\,ms in the census. The body's examples use max-autotune rather than reduce-overhead compilation. In Table~\ref{tab:formatinteractionfull}, share is bf16 prefill time divided by full bf16 inference time, speed-up is bf16 divided by int8 prefill time, and saving is the percentage of whole-inference time removed by int8.


\begin{table*}[t]
\centering
\footnotesize
\caption{Earlier A10G int8-compute measurements under reduce-overhead compilation, CUDA graphs and FlashAttention-2. These differ from the max-autotune examples in the body. Share is bf16 prefill time divided by total bf16 inference time. Speed-up is bf16 divided by int8 prefill time. Saving is the percentage of full inference time removed by int8 at that pruning level. These rows retain an earlier measurement job, rather than the independent later census jobs.}
\label{tab:formatinteractionfull}
\label{tab:formatinteraction}
\setlength{\tabcolsep}{8pt}
\begin{adjustbox}{max width=\textwidth}
\begin{tabular}{@{}lrrrrr@{}}
\toprule
Model & pruned (\%) & bf16 / int8 prefill (ms) & prefill share (\%) & prefill speed-up & inference saving (\%) \\
\midrule
OpenVLA-OFT & 0 & 141.2 / 83.6 & 85.0 & 1.69 & 34.6 \\
OpenVLA-OFT & 75 & 88.0 / 40.2 & 78.0 & 2.19 & 42.3 \\
CogACT & 0 & 113.8 / 40.2 & 81.1 & 2.83 & 52.4 \\
CogACT & 75 & 43.7 / 22.9 & 62.3 & 1.91 & 29.5 \\
\bottomrule
\end{tabular}
\end{adjustbox}
\end{table*}

The upper block of Table~\ref{tab:interactionabsolute} applies CUDA graphs before pruning (with compilation for $\pi_{0.5}$). The lower block applies quantization after pruning under reduce-overhead compilation.

\begin{table*}[p]
\centering
\footnotesize
\caption{Earlier four-latency comparisons, in milliseconds. $T_0$ is the baseline, $T_1$ and $T_2$ apply the first and second method alone, and $T_{12}$ applies both. The upper group adds CUDA graphs before pruning (with compilation for $\pi_{0.5}$), and the lower adds quantisation after pruning under reduce-overhead compilation. All rows remove 75\% of visual tokens. Savings are milliseconds removed by the second method before / after the first. FP8s means static FP8, int8c int8 compute, and int4w int4 weights. Bold marks $T_{12}$ values meeting $B$ before deployment reserves. Measurement provenance is explained in Appendix~\ref{app:extra}.}
\label{tab:interactionabsolute}
\setlength{\tabcolsep}{4pt}
\begin{adjustbox}{max width=\textwidth}
\begin{tabular}{lllrrrrrr}
\toprule
Model & platform & quant. & $T_0$ & $T_1$ & $T_2$ & $T_{12}$ & savings before / after & budget \\
\midrule
\multicolumn{9}{l}{\emph{Graphs before pruning}} \\
OpenVLA-OFT & H100 & -- & 42.44 & 33.94 & 31.86 & \textbf{22.74} & 10.58 / 11.19 & 150 \\
OpenVLA-OFT & A10G & -- & 204.96 & 187.67 & 138.57 & \textbf{121.26} & 66.39 / 66.41 & 150 \\
OpenVLA-OFT & Thor & -- & 180.57 & 176.66 & 113.21 & \textbf{109.33} & 67.36 / 67.34 & 150 \\
OpenVLA-OFT & RTX 4090 & -- & 101.78 & 77.87 & 67.90 & \textbf{44.03} & 33.88 / 33.85 & 150 \\
CogACT & H100 & -- & 50.75 & 27.79 & 47.30 & \textbf{23.92} & 3.44 / 3.87 & 50 \\
CogACT & A10G & -- & 202.95 & 148.73 & 129.04 & 74.93 & 73.91 / 73.80 & 50 \\
CogACT & Thor & -- & 134.43 & 111.77 & 119.61 & 97.68 & 14.81 / 14.09 & 50 \\
CogACT & RTX 4090 & -- & 106.59 & 50.53 & 96.16 & \textbf{36.11} & 10.43 / 14.42 & 50 \\
OpenVLA & H100 & -- & 103.81 & 75.02 & 98.31 & 68.80 & 5.51 / 6.22 & 50 \\
OpenVLA & A10G & -- & 349.07 & 332.36 & 273.79 & 255.00 & 75.28 / 77.36 & 50 \\
OpenVLA & Thor & -- & 444.80 & 444.60 & 429.73 & 426.33 & 15.07 / 18.28 & 50 \\
OpenVLA & RTX 4090 & -- & 210.63 & 142.40 & 200.02 & 126.08 & 10.61 / 16.32 & 50 \\
LLaVA-VLA & H100 & -- & 483.58 & 385.16 & 473.32 & 345.36 & 10.27 / 39.79 & 100 \\
LLaVA-VLA & A10G & -- & 1423.99 & 1394.45 & 1328.20 & 1282.66 & 95.79 / 111.79 & 100 \\
LLaVA-VLA & Thor & -- & 2253.42 & 2283.09 & 2155.74 & 2150.77 & 97.68 / 132.33 & 100 \\
LLaVA-VLA & RTX 4090 & -- & 998.24 & 695.45 & 966.19 & 635.89 & 32.05 / 59.56 & 100 \\
$\pi_{0.5}$ & H100 & -- & 111.63 & 33.99 & 113.51 & \textbf{30.69} & -1.88 / 3.30 & 200 \\
$\pi_{0.5}$ & A10G & -- & 323.80 & 116.61 & 309.34 & \textbf{94.06} & 14.46 / 22.55 & 200 \\
$\pi_{0.5}$ & Thor & -- & 207.73 & 116.11 & 199.19 & \textbf{104.01} & 8.54 / 12.10 & 200 \\
$\pi_{0.5}$ & RTX 4090 & -- & 249.99 & 55.49 & 250.75 & \textbf{46.23} & -0.75 / 9.26 & 200 \\
\multicolumn{9}{l}{\emph{Pruning before quantisation}} \\
OpenVLA-OFT & H100 & FP8s & 26.89 & 16.93 & 18.43 & \textbf{13.09} & 8.46 / 3.84 & 150 \\
OpenVLA-OFT & A10G & int8c & 166.02 & 112.87 & 108.58 & \textbf{65.13} & 57.44 / 47.74 & 150 \\
OpenVLA-OFT & Thor & FP8s & 152.56 & 92.37 & 90.79 & \textbf{62.68} & 61.76 / 29.69 & 150 \\
CogACT & H100 & FP8s & 21.42 & 17.51 & 18.37 & \textbf{15.94} & 3.06 / 1.57 & 50 \\
CogACT & A10G & int8c & 140.32 & 70.17 & 66.78 & \textbf{49.45} & 73.54 / 20.73 & 50 \\
CogACT & Thor & FP8s & 95.82 & 89.91 & 67.91 & 61.43 & 27.91 / 28.48 & 50 \\
OpenVLA & H100 & FP8s & 53.04 & 46.69 & 36.95 & \textbf{34.71} & 16.09 / 11.98 & 50 \\
OpenVLA & A10G & int4w & 311.57 & 238.33 & 143.28 & 102.13 & 168.28 / 136.20 & 50 \\
OpenVLA & Thor & int4w & 418.93 & 408.31 & 158.06 & 133.43 & 260.88 / 274.88 & 50 \\
LLaVA-VLA & H100 & FP8s & 273.09 & 235.87 & 187.58 & 169.10 & 85.51 / 66.77 & 100 \\
LLaVA-VLA & A10G & int4w & 1297.30 & 1200.99 & 528.91 & 407.19 & 768.39 / 793.80 & 100 \\
LLaVA-VLA & Thor & int4w & 2188.98 & 2086.57 & 740.27 & 653.36 & 1448.71 / 1433.21 & 100 \\
\bottomrule
\end{tabular}
\end{adjustbox}
\end{table*}

\paragraph{Compilation and memory costs.}
Panel (a) of Figure~\ref{fig:neutralapp} in the main paper distinguishes latency speed-up (bar height) from kernel-count reduction (bar label). Fusion reduces kernel counts by 1.9--2.8 times.

\begin{figure*}[p]
\centering
\includegraphics[width=\textwidth]{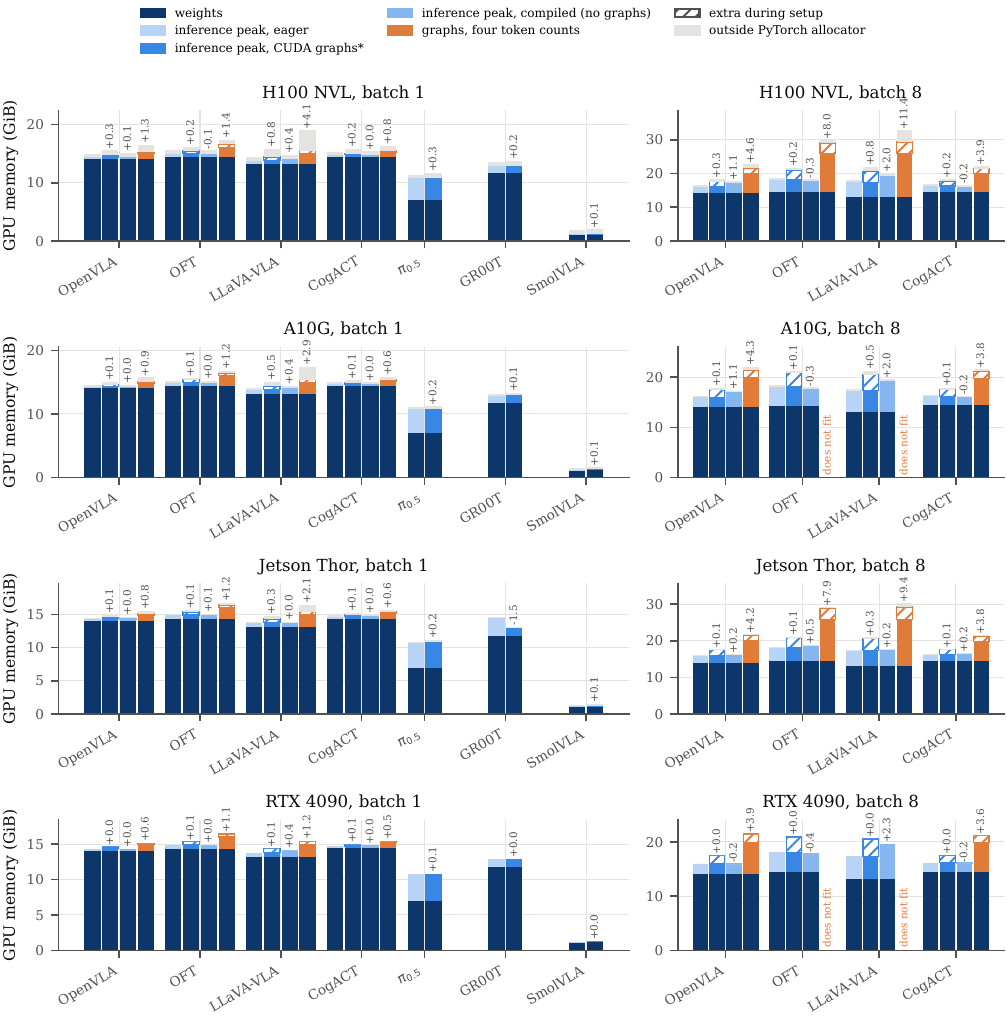}
\caption{GPU memory by model, platform and execution mode at batches one and eight. Stacks separate weights, inference memory, setup-only memory and the outside-allocator estimate. Signed labels give GiB above eager execution. An asterisk marks CUDA-graph modes that include compilation for flow models.}
\label{fig:memory}
\end{figure*}

Figure~\ref{fig:memory} expands the memory measurements by platform. Each batch size and retained-token count needs its own CUDA graph and reserved memory pool. Four pruning-level graphs held at once at batch eight exceed 24\,GB for OpenVLA-OFT and LLaVA-VLA. The hatched portions are additional allocator memory needed during setup; the gray portions estimate memory outside PyTorch's allocator from process usage. Signed labels show the memory difference from eager execution during inference. They include estimated memory outside PyTorch's allocator but exclude the temporary setup increment.

\paragraph{Denoising and other execution costs.}
Figure~\ref{fig:steps} shows how inference latency changes with denoising-step count. Compilation reduces the slope for $\pi_{0.5}$ and SmolVLA, leaving less time to save by removing steps. GR00T on Thor shows the reverse. Table~\ref{tab:costs} summarizes linear fits: the slope is time per step and the intercept is time outside the loop. The lines in the figure connect measurements rather than showing the fits.

\paragraph{Stage, selection and setup costs.}
Table~\ref{tab:costs} also reports Llama-2-7B prefill and decode costs under CUDA graphs. Its compiled decode row uses max-autotune on PyTorch 2.10 and torchao 0.13.

\begin{table}[t]
\centering
\scriptsize
\caption{Time costs in milliseconds for one request at a time. A straight line through latency at the measured denoising step counts gives time per step as its slope and time outside the loop as its intercept. Entries show eager / compiled execution with graphs. Llama-2-7B prefill time per 100 tokens and decoding time per token are measured under graphs. The compiled decode row is OpenVLA's decode time per generated token under max-autotune with PyTorch 2.10 and torchao 0.13 on each platform, unmeasured on the RTX 4090.}
\label{tab:costs}
\setlength{\tabcolsep}{3pt}
\begin{adjustbox}{max width=\columnwidth}
\begin{tabular}{@{}lrrrr@{}}
\toprule
 & H100 & A10G & Thor & RTX 4090 \\
\midrule
$\pi_{0.5}$, one step & 8.6 / 1.9 & 21 / 4.3 & 14 / 6.1 & 19 / 2.3 \\
\quad rest of the inference & 26 / 15 & 105 / 74 & 73 / 56 & 61 / 32 \\
SmolVLA, one step & 18 / 2.8 & 40 / 3.9 & 27 / 3.7 & 37 / 2.6 \\
\quad rest of the inference & 27 / 26 & 65 / 65 & 46 / 45 & 55 / 55 \\
GR00T N1.7, one step & 12 / 5.1 & 27 / 21 & 14 / 27 & 27 / 8.3 \\
\quad rest of the inference & 24 / 24 & 66 / 68 & 97 / 80 & 52 / 51 \\
\addlinespace[2pt]Llama-2-7B, 100 prefill tokens & 3.0 & 24 & 12 & 9.2 \\
Llama-2-7B, one decoded token & 9.6 & 33 & 59 & 17 \\
\quad under max-autotune compilation & 5.7 & 27 & 52 & \na \\
\bottomrule
\end{tabular}
\end{adjustbox}
\end{table}

Table~\ref{tab:selector} shows token-selection overhead. Standard FastV requires unfused attention to obtain attention weights; the fused scoring implementation retains the fused attention path.

\begin{table}[t]
\centering
\footnotesize
\caption{Token-selection overhead in eager execution. Entries are inference times in milliseconds, either unpruned or with 75\% of visual tokens removed at layer 3. FastV ranks tokens using attention weights, which the standard fused kernel does not write to memory, so this implementation uses unfused attention. A fused scoring implementation (fastv\_fa) retains the fused path.}
\label{tab:selector}
\begin{adjustbox}{max width=\columnwidth}
\begin{tabular}{@{}llrrrr@{}}
\toprule
Model & setting & H100 NVL & A10G & Thor & RTX 4090 \\
\midrule
OpenVLA-OFT & unpruned, fused attention & 41.8 & 204.1 & 180.6 & 102.1 \\
 & unpruned, unfused attention & 50.7 & 237.0 & 243.5 & 110.2 \\
 & random, fused attention & 30.6 & 137.1 & 113.3 & 68.3 \\
 & FastV, unfused attention & 33.2 & 143.8 & 124.8 & 70.0 \\
 & FastV scored by a fused kernel & 30.6 & 136.9 & 113.5 & 68.5 \\
\bottomrule
\end{tabular}
\end{adjustbox}
\end{table}

Table~\ref{tab:setup} reports startup from a loaded model through its first inference. We report the longest observed startup, but prior compiler-cache reuse can still shorten it. Eager startup takes 0.3--5.2 seconds.

\begin{table}[H]
\centering
\scriptsize
\caption{Time from an already loaded model to completion of the first inference, in seconds, for one bf16 request. Columns show CUDA graph setup (graphs), compilation with CUDA graphs (comp.), and max-autotune with CUDA graphs (m.-a.). Flow-model runtimes combine CUDA graphs and compilation. Each entry gives the longest observed startup because cached compiler results may shorten later runs. A dagger marks a preceding compilation of the same model on the same host, so startup without cached results could be longer. RTX 4090 used a different compilation mode and is omitted. Eager startup takes 0.3--5.2 seconds.}
\label{tab:setup}
\setlength{\tabcolsep}{3pt}
\begin{adjustbox}{max width=\columnwidth}
\begin{tabular}{@{}lrrrrrrrrr@{}}
\toprule
Model & \multicolumn{3}{c}{H100} & \multicolumn{3}{c}{A10G} & \multicolumn{3}{c}{Thor} \\
 & graphs & comp. & m.-a. & graphs & comp. & m.-a. & graphs & comp. & m.-a. \\
\midrule
OpenVLA & 1.7 & 31 & 41 & 4.7 & 54 & 73 & 2.0 & 48 & 68 \\
OpenVLA-OFT & 2.2 & 40 & 58 & 3.4 & 69 & 101 & 0.8 & 67 & 94 \\
LLaVA-VLA & 4.2 & 37 & 52 & 8.0 & 65 & 89 & 10 & 63 & 92 \\
CogACT & 0.8 & 6.8 & 43 & 1.2 & 52 & 74 & 0.6 & 49 & 67 \\
$\pi_{0.5}$ & na & 80 & 139 & \na & 120 & 216 & \na & 29$^\dagger$ & 181 \\
GR00T N1.7 & \na & 8.3$^\dagger$ & \na & \na & 93 & \na & \na & 105 & \na \\
SmolVLA & \na & 7.9$^\dagger$ & \na & \na & 10$^\dagger$ & \na & \na & 39 & \na \\
\bottomrule
\end{tabular}
\end{adjustbox}
\end{table}

Table~\ref{tab:platforms} documents hardware and software. Its eager inference time per kernel divides total unprofiled time by kernel count for models with estimated time outside kernels above 40\%. This is not a direct CPU-time measurement.

\begin{table*}[t]
\centering
\footnotesize
\caption{Platforms, software and measured reference parameters. Bandwidth here is measured throughput. The weight-read exclusion rule instead uses rated datasheet bandwidth. The ridge point $C/BW$ is the ideal compute/memory crossover in FLOP/byte. When bf16 weight reads dominate, it also approximates the token-count crossover. $\tau_{\rm dev}$ is minimum kernel duration and $\tau_{\rm host}$ the CPU microbenchmark time per launch. Preprocessing entries are medians across models with ranges in brackets.}
\label{tab:platforms}
\label{tab:devices}
\label{tab:versions}
\label{tab:launchcost}
\begin{adjustbox}{max width=\textwidth}
\begin{tabular}{@{}lrrrr@{}}
\toprule
Platform Property & H100 NVL & A10G & Jetson Thor & RTX 4090 \\
\midrule
\multicolumn{5}{@{}l}{\emph{Hardware and Measured Roofs}} \\
Host processor & Xeon Gold 6548N & EPYC 7R32 & Arm Neoverse-V3AE & EPYC 7543 \\
bf16 GEMM peak $C$ (TFLOP/s) & 557 & 67 & 153 & 156 \\
DRAM bandwidth $BW$ (GB/s) & 3650 & 519 & 253 & 955 \\
Ridge point $C/BW$ (FLOP/B) & 152 & 129 & 605 & 163 \\
Device minimal kernel $\tau_{\rm dev}$ ($\mu$s) & 1.4 & 1.0 & 1.0 & 0.8 \\
Host dispatch microbenchmark $\tau_{\rm host}$ ($\mu$s/op) & 5.2 & 6.8 & 4.7 & 4.7 \\
Eager inference time per kernel ($\mu$s) & 8.5 to 11.2 & 18.2 to 22.4 & 28.5 to 35.1 & 19.8 to 24.3 \\
\addlinespace[2pt]\multicolumn{5}{@{}l}{\emph{Timing Boundary Reserve}} \\
Preprocessing reserve $t_{\rm preproc}$ (median [range], ms) & 4.0 [0.8--6.1] & 5.7 [1.1--9.2] & 4.4 [1.4--9.2] & 6.1 [1.0--8.9] \\
\addlinespace[2pt]\multicolumn{5}{@{}l}{\emph{Software Environment}} \\
PyTorch & 2.7.1 / 2.10.0 / 2.11.0 & 2.7.1 / 2.10.0 / 2.11.0 & 2.10.0 (JetPack R39) & 2.7.1 / 2.11.0 \\
CUDA & 12.6 & 12.6 & 13.0 & 12.6 / 12.8 \\
torchao & 0.12.0 / 0.13.0 & 0.12.0 / 0.13.0 & 0.13.0 & 0.12.0 \\
FlashAttention & 2.8.3 & 2.8.3 & 2.8.4 & 2.8.3 \\
\bottomrule
\end{tabular}
\end{adjustbox}
\end{table*}

\paragraph{Quantization implementations.}
Table~\ref{tab:formats} compares quantization with bf16 under reduce-overhead compilation. Weight-only schemes retain bf16 activations. Dynamic FP8 scales activations during inference; static FP8 uses scales calibrated beforehand. FP8 needs hardware support absent on A10G, and NVFP4 is evaluated on Thor. In the table, action denotes OpenVLA's decode loop or OpenVLA-OFT's action head.

\begin{table}[t]
\centering
\scriptsize
\caption{Quantisation speed-ups over compiled bf16 with CUDA graphs, at batch one without pruning, using the fastest selected library per GPU. Action means OpenVLA's decode loop or OpenVLA-OFT's action head. Values below one indicate slowdowns. Bold marks the best quantisation setting in a model's column when its speed-up is at least 5\%. Dynamic FP8 scales activations per inference. Static FP8 uses scales calibrated in advance. NVFP4 requires FP4 tensor cores (Thor), and FP8 requires FP8 tensor cores (unavailable on the A10G).}
\label{tab:formats}
\setlength{\tabcolsep}{2.4pt}
\begin{adjustbox}{max width=\columnwidth}
\begin{tabular}{@{}llrr@{\hspace{6pt}}rr@{\hspace{6pt}}rr@{}}
\toprule
Model & quantisation & \multicolumn{2}{c}{H100} & \multicolumn{2}{c}{A10G} & \multicolumn{2}{c}{Thor} \\
 & & prefill & action & prefill & action & prefill & action \\
\midrule
OpenVLA-OFT & int8 weights & 0.82 & 1.00 & 0.93 & 1.00 & 1.33 & 1.00 \\
 & int4 weights & 0.89 & 1.01 & 1.00 & 0.99 & 1.62 & 1.02 \\
 & int8 compute & 1.24 & 1.03 & \textbf{1.69} & 1.00 & 1.88 & 1.00 \\
 & FP8 dynamic & 1.21 & 1.02 & \na & \na & 1.77 & 1.01 \\
 & FP8 static & \textbf{1.68} & 1.02 & \na & \na & \textbf{1.95} & 1.02 \\
 & NVFP4 compute & \na & \na & \na & \na & 1.49 & 1.01 \\
\addlinespace[2pt]
OpenVLA & int8 weights & 0.68 & 1.39 & 1.53 & 2.13 & 1.15 & 1.97 \\
 & int4 weights & 0.74 & \textbf{1.64} & \textbf{1.54} & \textbf{3.28} & 1.53 & \textbf{3.33} \\
 & FP8 dynamic & 1.26 & 1.45 & \na & \na & 1.45 & 1.96 \\
 & FP8 static & \textbf{1.36} & 1.53 & \na & \na & \textbf{1.57} & 1.95 \\
 & NVFP4 compute & \na & \na & \na & \na & 1.43 & 2.60 \\
\addlinespace[2pt]
\bottomrule
\end{tabular}
\end{adjustbox}
\end{table}

Table~\ref{tab:survey} lists the implementations selected using the speed and error criterion in Appendix~\ref{app:runtimes}. The survey averages prefill results at two token counts, 291 and 605. Its speed-up need not match the whole-prefill result for an individual model.

Gemlite is a Triton kernel library~\cite{gemlite} and tinygemm an int4 kernel, both accessed through torchao~\cite{torchao}. Marlin is the mixed-precision kernel of \citet{frantar2025marlin}. Machete and vLLM CUTLASS are vLLM's mixed-input and int8 kernels. CUTLASS denotes NVIDIA's template kernels~\cite{cutlasslibrary}, and FlashInfer supplies the NVFP4 kernels~\cite{flashinferlibrary}.

\begin{table*}[t]
\centering
\footnotesize
\caption{Selected implementations and their speed-ups over bf16, shown in parentheses. The survey times the seven linear layers within one compiled Llama-2-7B decoder layer, using PyTorch 2.10 and torchao 0.13. Decode uses one token; prefill averages results at 291 and 605 tokens. For each stage and quantisation setting, we select the fastest implementation whose output error is within 1.25 times the smallest measured error. A speed-up below one means no tested implementation beats bf16. Appendix~\ref{app:extra} describes the libraries and the difference from whole-prefill measurements.}
\label{tab:survey}
\setlength{\tabcolsep}{4pt}
\begin{adjustbox}{max width=\textwidth}
\begin{tabular}{@{}lll@{\hspace{8pt}}ll@{\hspace{8pt}}ll@{}}
\toprule
quantisation & \multicolumn{2}{c}{H100} & \multicolumn{2}{c}{A10G} & \multicolumn{2}{c}{Thor} \\
 & decode & prefill & decode & prefill & decode & prefill \\
\midrule
int8 weights & Gemlite (1.35) & Machete (0.79) & Gemlite (2.13) & Gemlite (1.17) & Marlin (2.00) & CUTLASS (1.31) \\
int4 weights & torchao tinygemm (1.59) & Machete (0.86) & torchao tinygemm (3.26) & Marlin (1.22) & torchao tinygemm (3.41) & CUTLASS (1.65) \\
int8 compute & Gemlite (1.39) & vLLM CUTLASS (1.26) & Gemlite (2.04) & Gemlite (2.15) & Gemlite (1.99) & torchao (1.73) \\
FP8 dynamic & Gemlite (1.39) & Gemlite (1.24) & \na & \na & Gemlite (2.03) & torchao (1.65) \\
FP8 static & torchao (1.49) & torchao (1.55) & \na & \na & torchao (1.97) & torchao (1.78) \\
NVFP4 & \na & \na & \na & \na & FlashInfer (2.64) & FlashInfer (1.51) \\
\bottomrule
\end{tabular}
\end{adjustbox}
\end{table*}

\section{Platform Notes}
\label{app:platformnotes}

\begin{figure*}[t]
\centering
\includegraphics[width=\textwidth]{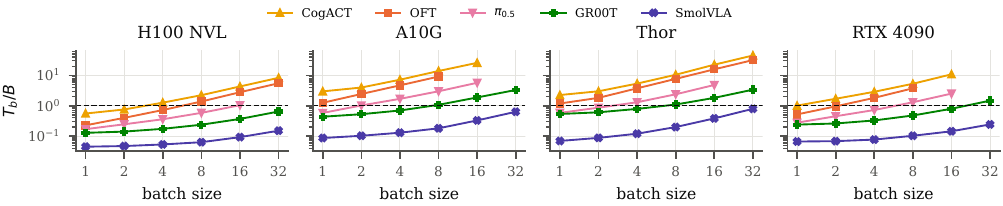}
\caption{Unpruned batch latency $T_b$ relative to the nominal inference budget $B$. Colors and markers identify models; both axes are logarithmic. The dashed line marks $T_b=B$. Curves use the execution modes stated in the text and do not estimate robot capacity.}
\label{fig:batching}
\end{figure*}

\paragraph{Serving multiple robots.}
After finding a combination that meets the budget, a server can test whether the remaining time permits processing more requests together.
Figure~\ref{fig:batching} first checks whether the entire batch finishes within the nominal inference budget. CogACT and OpenVLA-OFT use CUDA graphs; the flow models use reduce-overhead compilation with graphs. Each curve uses one mode per model, with no pruning. This batch-latency check does not determine robot capacity, because serving also depends on request arrival times and queue scheduling.
Table~\ref{tab:fleet} therefore estimates serving capacity by combining the measured batch latencies with the asynchronous scheduler replay. The bf16 serving replay dynamically batches waiting observations from robots with fixed random phase offsets. The reported robot counts apply to this request pattern and the replay's scheduling assumptions, and its 5\,ms server ticks make results close to the budget sensitive to scheduling.
On the Thor, over-current protection events were observed during OpenVLA-OFT graph runs at batches 2, 4 and 8 and during compiled $\pi_{0.5}$ runs at batches 4 and 16 but not 8. Repeated OpenVLA-OFT runs agreed within 2\%, so the counts are reported.

\begin{table*}[t]
\centering
\footnotesize
\caption{Number of robots supported by measured batch times and the action-queue simulation at 20\,Hz. Each entry gives the maximum robot count without batching / with the best tested batch-size cap, allowing each robot to wait for actions on at most 1\% of control steps. Budgets are in parentheses. Zero means even one robot fails this check. Thor protection events occurred for OFT graphs at batches 2, 4 and 8 and compiled $\pi_{0.5}$ at batches 4 and 16, but not at batch 8. OFT repeat latencies agreed within 2\%. Near-budget results depend on the replay's scheduling assumptions (Appendix~\ref{app:platformnotes}).}
\label{tab:fleet}
\begin{adjustbox}{max width=\textwidth}
\begin{tabular}{llrrrr}
\toprule
Model & mode & H100 NVL & A10G & Jetson Thor & RTX 4090 \\
\midrule
OpenVLA-OFT (150\,ms) & eager & 3 / 4 & 0 / 0 & 0 / 0 & 1 / 1 \\
 & eager, 75\% pruned & 4 / 8 & 1 / 1 & 1 / 2 & 1 / 2 \\
 & CUDA graphs & 3 / 4 & 0 / 0 & 0 / 0 & 1 / 2 \\
 & CUDA graphs, 75\% pruned & 5 / 8 & 1 / 1 & 1 / 2 & 2 / 2 \\
$\pi_{0.5}$ (200\,ms) & eager & 1 / 8 & 0 / 0 & 1 / 1 & 0 / 0 \\
 & eager, 75\% pruned & 1 / 16 & 0 / 0 & 1 / 2 & 0 / 0 \\
 & compiled & 5 / 9 & 1 / 2 & 1 / 2 & 3 / 4 \\
 & compiled, 75\% pruned & 5 / 17 & 1 / 2 & 1 / 4 & 3 / 8 \\
GR00T (350\,ms) & eager & 3 / 16 & 1 / 4 & 2 / 4 & 1 / 8 \\
 & compiled & 5 / 16 & 2 / 4 & 1 / 4 & 3 / 8 \\
SmolVLA (1200\,ms) & eager & 4 / 80 & 2 / 32 & 3 / 16 & 2 / 51 \\
 & compiled & 12 / 112 & 6 / 33 & 8 / 32 & 8 / 83 \\
\bottomrule
\end{tabular}
\end{adjustbox}
\end{table*}

\paragraph{The Thor's power modes.}
Power-mode measurements vary CPU and GPU resources on one device (Table~\ref{tab:power-thor-agx-128gb}). On the Thor, both 90 and 70 W modes disable 8 of 20 streaming multiprocessors. Of the lower modes only 70 W lowers the CPU clock, from 2601 to 1998\,MHz, according to recorded clocks. The other modes reduce GPU resources without lowering the CPU clock.
Disabling streaming multiprocessors mainly slows prefill-heavy models, while leaving the decoding models almost unchanged, consistent with their weight-read costs.
In these measurements, lowering the CPU clock mainly slows cases where the GPU waits for CPU launches. The compiled GR00T and SmolVLA entries at MAXN come from the denoising-step sweep, which was measured only without pruning. The $\pi_{0.5}$ graph mode includes compilation.

\begin{table*}[t]
\centering
\footnotesize
\caption{Jetson Thor power-mode measurements. Latency entries show unpruned / 75\% visual-token pruning, in milliseconds. Energy is joules per inference without pruning, measured on the GPU and CPU rails. MAXN is the default baseline in Section~\ref{sec:method}. $\pi_{0.5}$'s graph mode includes compilation. Compiled GR00T and SmolVLA at MAXN use the full-step denoising runs, available only without pruning. }
\label{tab:power-thor-agx-128gb}
\begin{adjustbox}{max width=\textwidth}
\begin{tabular}{llrrrrrrrr}
\toprule
 & & \multicolumn{2}{c}{MAXN} & \multicolumn{2}{c}{120 W} & \multicolumn{2}{c}{90 W} & \multicolumn{2}{c}{70 W} \\
Model & exec. & ms & J/infer. & ms & J/infer. & ms & J/infer. & ms & J/infer. \\
\midrule
OpenVLA & eager & 445/430 & 20.93 & 453/435 & 20.46 & 455/429 & 19.23 & 453/428 & 18.74 \\
OpenVLA & graph & 445/426 & 20.22 & 450/430 & 19.95 & 453/425 & 18.93 & 453/425 & 18.52 \\
OpenVLA-OFT & eager & 181/113 & 14.32 & 186/117 & 14.26 & 212/125 & 13.71 & 213/126 & 13.51 \\
OpenVLA-OFT & graph & 177/109 & 14.28 & 182/113 & 14.10 & 209/122 & 13.42 & 209/123 & 13.37 \\
LLaVA-VLA & eager & 2253/2156 & 100.47 & 2303/2203 & 100.14 & 2273/2154 & 89.31 & 2285/2160 & 87.47 \\
LLaVA-VLA & graph & 2283/2151 & 96.69 & 2320/2202 & 96.50 & 2282/2146 & 88.22 & 2306/2158 & 86.89 \\
CogACT & eager & 134/120 & 7.96 & 137/121 & 7.89 & 147/128 & 7.89 & 154/133 & 7.79 \\
CogACT & graph & 112/98 & 7.47 & 116/101 & 7.38 & 124/104 & 7.38 & 125/104 & 7.26 \\
$\pi_{0.5}$ & eager & 208/199 & 9.05 & 220/210 & 9.20 & 237/216 & 9.31 & 268/248 & 9.49 \\
$\pi_{0.5}$ & graph & 116/104 & 6.59 & 119/106 & 6.57 & 135/111 & 6.75 & 135/112 & 6.68 \\
GR00T N1.7 & eager & 149/147 & 6.90 & 155/152 & 6.90 & 162/158 & 6.63 & 170/166 & 6.52 \\
GR00T N1.7 & compiled & 188/\na & 8.80 & 195/192 & 8.94 & 208/203 & 8.66 & 210/205 & 8.51 \\
SmolVLA & eager & 308/308 & 7.35 & 312/317 & 7.30 & 323/324 & 6.67 & 372/371 & 7.02 \\
SmolVLA & compiled & 82/\na & 2.87 & 86/83 & 2.93 & 94/89 & 2.80 & 100/94 & 2.81 \\
\bottomrule
\end{tabular}
\end{adjustbox}
\end{table*}

\paragraph{GR00T compiled on the Thor.}\label{app:groot}
Compilation slows GR00T on the Thor. Between the eager and compiled profiles, gaps between kernels shrink from 42.1 to 18.7\,ms, while the time covered by kernels grows from 134.8 to 178.8\,ms.
Among the most frequent kernels in each trace, those whose names identify them as copies or data-type conversions take 70.4\,ms per inference in eager execution and 136.8\,ms when compiled, and 93.9\,ms of the compiled figure is in compiler-generated Triton kernels.
These are profiler-session measurements, not an additive decomposition of the separate unprofiled latencies in Table~\ref{tab:power-thor-agx-128gb}.
The slowdown also appears with pruning and in every measured power mode (Table~\ref{tab:power-thor-agx-128gb}).

\section{Literature Audit}
\label{app:audit}
We audited 25 VLA acceleration papers that report inference speed-ups, token pruning, caching, or quantization (Table~\ref{tab:litaudit}).
The entries are the papers we identified that report a latency or speed-up for one of these methods on a named VLA, checked against each paper's arXiv text. The list is not the result of a systematic database search and is not claimed to be exhaustive.
Of the 13 entries listing token pruning, reuse, merging or sparsification, none specifies the baseline and method execution modes. One names only its attention backend.
Among the step-reduction papers, \citet{tang2026fewersteps} report inference without compilation, using SDPA attention.
For papers omitting execution-mode information, readers cannot check whether a reported saving applies to compiled execution or CUDA graphs.

\begin{table*}[t]
\centering
\small
\caption{How VLA speed-up papers measure latency. GPU: the device behind the paper's latency
numbers (``--'': none named). Mode: whether the paper says if its baseline ran eagerly, compiled,
as CUDA graphs or under TensorRT. $\dagger$: baseline latency taken from another paper or measured
on a different GPU than the method.}
\label{tab:litaudit}
\begin{adjustbox}{max width=\textwidth}
\begin{tabular}{@{}llllp{3.9cm}@{}}
\toprule
Paper & Method & GPU & Mode & Baseline \\
\midrule
VLA-Cache~\cite{vlacache} & token reuse & RTX 4090 & no (bf16 only) & OpenVLA, OFT, CogACT \\
EfficientVLA~\cite{efficientvla} & pruning, caching & A40 & no & CogACT \\
SpecPrune-VLA~\cite{specprune} & pruning & A800, 3090, 4090 & attention only & OFT, eager and FlashAttention-2 attention \\
SP-VLA~\cite{spvla} & pruning, scheduling & RTX 4090 & no & OpenVLA, CogACT \\
FlashVLA~\cite{flashvla} & pruning, action reuse & H100 & no & OpenVLA \\
LightVLA~\cite{lightvla} & pruning & H20 & no & OFT, reproduced; other methods' rows on 4090/H100/A100$\dagger$ \\
VLA-Pruner~\cite{vlapruner} & pruning & RTX 4090 & no & OpenVLA, OFT, $\pi_0$ \\
ADP~\cite{adpvla} & pruning & RTX 4090 & no & OFT, public scripts \\
CogVLA~\cite{cogvla} & sparsification & A800 & no & OpenVLA, OFT, PD-VLA, reproduced \\
TEAM-VLA~\cite{teamvla} & token merging & A100 & no & OFT, from SpecPrune-VLA's A800 runs$\dagger$ \\
DepthCache~\cite{depthcache} & token merging & RTX 4090 & no & $\pi_{0.5}$, OpenVLA, GR00T; SP-VLA cited$\dagger$ \\
IAprune~\cite{iaprune} & pruning & A100 & no & OFT, DreamVLA, $\pi_0$, $\pi_{0.5}$ \\
\midrule
QAIL~\cite{qail} & quantisation & AGX Orin & method only & OpenVLA bf16 \\
BitVLA~\cite{bitvla} & 1-bit weights & A100 & no & OFT, from the OFT paper$\dagger$ \\
SQAP-VLA~\cite{sqapvla} & quantisation, pruning & RTX 3090 & no & CogACT fp16 \\
DeeR-VLA~\cite{deervla} & early exit & V100 & no & RoboFlamingo, reproduced \\
MoLe-VLA~\cite{molevla} & layer skipping & RTX 4090D & no & CogACT \\
PD-VLA~\cite{pdvla} & parallel decoding & -- & no & LLaVA-VLA \\
Spec-VLA~\cite{specvla} & speculative decoding & -- & no & OpenVLA \\
OpenVLA-OFT~\cite{openvlaoft} & parallel decoding & A100 & no & OpenVLA \\
TinyVLA~\cite{tinyvla} & smaller model & A6000 & no & OpenVLA \\
SmolVLA~\cite{shukor2025smolvla} & small model, async & -- & no & SmolVLA, synchronous \\
VLA-Adapter~\cite{vlaadapter} & small model & -- & no & OpenVLA, OFT (the OFT paper's A100 values)$\dagger$ \\
SnapFlow~\cite{snapflow} & fewer denoising steps & A800 & no & $\pi_{0.5}$, 10 steps \\
Fewer Steps~\cite{tang2026fewersteps} & fewer steps, correction & A100, 4090 & no compile; SDPA & $\pi_{0.5}$, SmolVLA \\
\bottomrule
\end{tabular}
\end{adjustbox}
\end{table*}

\section{CUDA Graphs: Changes in Kernel Time and Time Outside Kernels}
\label{app:closure}

Section~\ref{sec:stacking} tests whether percentage savings carry over. Here we examine changes in pruning's millisecond saving under CUDA graphs, which separate exactly into changes in kernel time and time outside kernels. Here $\Delta X$ means the value of $X$ before pruning minus its value afterwards. The difference between measured latency under CUDA graphs and the latency implied by unchanged millisecond savings is:
\begin{equation}
\begin{split}
\text{excess} &= T_{12} - (T_1 + T_2 - T_0) \\
&= (\Delta K_{\rm eager} - \Delta K_{\rm graph}) + (\Delta I_{\rm eager} - \Delta I_{\rm graph}),
\end{split}
\end{equation}
where $K$ is the total time covered by at least one GPU kernel in the CUPTI trace, $I = T - K$ is estimated time outside kernels, and $T$ is whole-inference latency.
The identity separates the measured difference into changes in kernel time, eager time outside kernels, and time outside kernels under CUDA graphs.
Positive excess means fewer milliseconds saved than assumed; negative excess means more. Only the change in time outside kernels under CUDA graphs enters the identity; its unpruned value alone does not determine the excess.

For OpenVLA, OpenVLA-OFT and CogACT on the H100 and A10G, pruning lowers estimated eager time outside kernels in 38 of 48 comparisons, by at most 0.72\,ms. The contribution from time outside kernels under CUDA graphs, $-\Delta I_{\rm graph}$, ranges from $-0.56$ to $+0.76$\,ms. The kernel-time contribution, $\Delta K_{\rm eager}-\Delta K_{\rm graph}$, ranges from $-2.0$ to $+0.08$\,ms.

The largest positive excess for an individual selector is 1.04\,ms (3.5\%, OpenVLA-OFT H100 at 25\% pruning). Its contributions include a 0.33\,ms reduction in eager time outside kernels and a 0.76\,ms increase in time outside kernels under CUDA graphs. After averaging paired selectors, the maximum millisecond-reuse error is 3.1\% of measured whole-inference latency (Table~\ref{tab:census_reuse}).

On the Thor, kernels keep the GPU busy for almost the whole inference. The kernel-time contribution ranges from $-10.7$ to $+11.1$\,ms. The contributions largely cancel, leaving a maximum positive excess of 2.5\,ms.
Estimated time outside kernels is 5.6\% to 6.9\% of unpruned latency under CUDA graphs on the H100 (1.6 to 5.2\,ms for OpenVLA-OFT, CogACT and OpenVLA; 25.7\,ms for LLaVA-VLA), compared with 0.1\% to 0.6\% on the A10G, whose longer kernels more easily hide CPU launches.
LLaVA-VLA's kernel term is $-21$\,ms and its excess $-28$\,ms on the H100, and $-19$ and $-35$\,ms on the Thor, because decoding under CUDA graphs attends over a fixed-size, masked key-value cache (Section~\ref{sec:principle}).

\section{Full Reuse Census Tables}
\label{app:census}

\paragraph{Comparisons and counting units.}
We call the full set of four-latency comparisons the reuse census. Each method-pair comparison records four times: the baseline $T_0$, each method alone ($T_1$, $T_2$), and both together ($T_{12}$).
The 714 comparisons comprise 682 for graphs, compilation and quantization on the four Llama-based models and 32 for step reduction on the flow models.
A \emph{configuration} fixes the model, platform, execution mode and optimization setting. Comparisons retain individual token selectors. Configuration-level results average paired selectors' four latencies before applying the reuse assumptions. Software and power variants count separately.

\paragraph{Matched compilation-and-pruning example.}
Table~\ref{tab:hardwareinteraction} gives the cross-platform inference times with matched software discussed in Section~\ref{sec:stacking}.
The eager and compiled prefill-only savings are 3.6 and 3.9\,ms on H100, versus 73.8 and 32.6\,ms on A10G.
These stage timings locate most of the difference in prefill without isolating a kernel cause.

\begin{table}[t]
\centering
\footnotesize
\caption{Pruning's whole-inference saving before and after compilation for CogACT. RO denotes reduce-overhead and MA max-autotune. Both GPUs use matching software versions and remove 75\% of visual tokens. All values are milliseconds. Full / pruned columns give latency without / with pruning in eager and compiled execution. Pruned times average the two token selectors.}
\label{tab:hardwareinteraction}
\setlength{\tabcolsep}{3pt}
\begin{adjustbox}{max width=\columnwidth}
\begin{tabular}{llrrrr}
\toprule
GPU & Mode & Eager full / pruned & Eager saving & Compiled saving & Compiled full / pruned \\
\midrule
H100 & RO & 49.9 / 46.0 & 3.9 & 4.0 & 21.6 / 17.6 \\
H100 & MA & 49.9 / 46.0 & 3.9 & 4.1 & 21.5 / 17.4 \\
A10G & RO & 200.1 / 125.9 & 74.2 & 70.1 & 140.2 / 70.1 \\
A10G & MA & 200.1 / 125.9 & 74.2 & 33.5 & 97.5 / 64.1 \\
\bottomrule
\end{tabular}
\end{adjustbox}
\end{table}

\paragraph{Reuse assumptions and budget decisions.}
Table~\ref{tab:census_reuse} compares measured latency with the latency implied by the reuse assumptions across all 714 comparisons. Percentage reuse reports $T_{\rm meas}/T_{\rm assumed}$; millisecond reuse reports $(T_{\rm meas}-T_{\rm assumed})/T_{\rm meas}$ where the implied latency is positive. Both count discrepancies using $T_{\rm meas}>1.053T_{\rm assumed}$; zero or negative implied latencies are counted separately. In the cutoff columns, parenthesized counts give distinct platform--model labels, retaining software and power variants. Counts describe coverage, not independent trials.
Table~\ref{tab:census_flips} provides the complete breakdown of budget decision flips across all control periods.
The 1.4\% band excludes small boundary differences and describes eager cross-job variation. A false accept has implied latency at or below $B$ but measured latency above $1.014B$; a false reject has implied latency above $B$ but measured latency below $0.986B$. In Table~\ref{tab:census_flips}, the CUDA graphs row counts software and power variants as one group, and the other rows count them separately. Neither this band nor the 5.3\% threshold is a calibrated bound.
Percentage reuse gives 12 false accepts and millisecond reuse 31, all in the Llama-based comparisons. Seven of the 12 are OpenVLA-OFT on Thor after pruning and quantization.

\begin{table*}[t]
\centering
\footnotesize
\caption{Latency implied by reuse assumptions versus measured combinations. Percentage reuse assumes the same percentage saving, and its ratio is measured latency divided by implied latency, $T_{\rm meas}/T_{\rm assumed}$. Millisecond reuse assumes the same time saving, and its error is $(T_{\rm meas}-T_{\rm assumed})/T_{\rm meas}$ for positive implied latencies. Counts above 5.3\% use $T_{\rm meas}>1.053T_{\rm assumed}$. Non-pos. counts zero or negative implied latencies. Comparisons retain individual token selectors; configurations average paired latencies. Pairs counts distinct platform and model combinations, retaining software and power variants. Counts describe coverage, not independent trials.}
\label{tab:census_reuse}
\begin{adjustbox}{max width=\textwidth}
\begin{tabular}{lrrrrrr}
\toprule
Step & Configs (comparisons) & Ratio: max / p95 & Above 5.3\% (pairs) & Error: max / p95 & Above 5.3\% (pairs) & Non-pos. \\
\midrule
CUDA graphs & 98 (134) & 1.006 / 0.999 & 0 (0) & 0.031 / 0.009 & 0 (0) & 0 \\
Compilation & 84 (156) & 1.188 / 1.087 & 8 (3) & 0.636 / 0.218 & 32 (8) & 0 \\
Pruning then quant., compiled & 194 (378) & 1.492 / 1.320 & 52 (5) & 0.813 / 0.454 & 83 (7) & 2 \\
Pruning then quant., graphs only & 14 (14) & 1.427 / 1.346 & 4 (3) & 0.980 / 0.430 & 5 (3) & 0 \\
Compilation, fewer steps (flow) & 32 (32) & 3.266 / 2.717 & 29 (11) & 0.968 / 0.947 & 9 (5) & 21 \\
\bottomrule
\end{tabular}
\end{adjustbox}

\end{table*}

\begin{table*}[t]
\centering
\footnotesize
\caption{Budget decisions under the reuse assumptions at control periods of 33.3, 50 and 100\,ms. A false accept occurs when the implied latency meets $B$ but measured latency exceeds $1.014B$. A false reject occurs when the implied latency exceeds $B$ but measured latency is below $0.986B$. Only the CUDA graphs row collapses software and power variants in its group count. Other rows retain those labels. Near $B$ means measured latency is within 20\% of the budget.}
\label{tab:census_flips}
\begin{tabular}{lrrr}
\toprule
Step & Decisions (groups; near $B$) & ms: false accept/reject & \%: false accept/reject \\
\midrule
CUDA graphs, pooled & 294 (183; 28) & 0 / 0 & 0 / 4 \\
Compilation & 252 (252; 40) & 6 / 2 & 0 / 7 \\
Pruning then quant., compiled & 582 (582; 55) & 23 / 4 & 10 / 5 \\
Pruning then quant., graphs only & 42 (42; 5) & 2 / 0 & 2 / 0 \\
Compilation, fewer steps (flow) & 96 (96; 0) & 0 / 0 & 0 / 0 \\
\bottomrule
\end{tabular}

\end{table*}

For the flow models, the baseline uses the runtime's default step count without pruning. We compare reduced step counts under reduce-overhead compilation.
The 32 flow combinations meet all three tested budgets, so their latency discrepancies change none of the 96 budget decisions.

\paragraph{Changes in quantization saving.}
The change in quantization saving is split into a contribution from the affected-stage share, $(\sigma_1-\sigma_0)(1-1/s_0)$, and one from the stage speed-up, $\sigma_1(1/s_0-1/s_1)$. Subscripts 0 and 1 mean before and after pruning.
In the 103 individual-selector comparisons from the H100, A10G and Thor campaigns above the 5.3\% threshold of Section~\ref{sec:stacking}, the affected-stage share falls in all 103 and the stage speed-up also falls in 99.
Table~\ref{tab:census_reuse} pools these campaigns with earlier measurements, which add none above the threshold, so 52 configurations exceed it after selector averaging.
Updating the share alone leaves 72 of the 368 individual-selector quantization comparisons across all campaigns above the threshold. The largest ratio of measured latency to latency implied by the updated assumption is 1.385.

The share and speed-up decomposition omits changes outside the affected prefill/decode stages. The absolute difference between attributed and measured whole-inference saving has a median of 0.4 percentage points and a maximum of 4.1 points. The attribution is therefore approximate.

\section{Preference Order and Screening Consistency}
\label{app:screeningobjective}
Table~\ref{tab:triage} lists the recorded settings that meet the inference-time budget for each over-budget model--platform case, using a fixed preference order. For pruning, select the lowest percentage of tokens removed that meets the budget. For quantization, consider groups in this order: int8 weights, int4 weights, 8-bit compute (int8, dynamic FP8 or static FP8), then NVFP4. Select the fastest scheme in the first group containing a budget-meeting option. Consider pairs only when neither method meets the budget alone, ordered by pruning level, then quantization group, then latency.
This order is fixed solely to make the comparison on the recorded grid reproducible. It does not rank task accuracy or prescribe a quantization order for deployment.
Each case uses one execution mode and only the settings recorded in it. Pruning uses random token selection; recorded levels vary by platform, with 6.25\% increments on RTX 4090. H100 and A10G cases use the PyTorch 2.10 and torchao 0.13 campaign, while LLaVA-VLA compilation and the other platforms retain earlier records.

Checking the screens of Section~\ref{sec:procedure} against the recorded measurements confirms that neither excludes a budget-meeting candidate.
The all-tokens-removed latency used by the affected-stage screen is recorded only in eager execution and under CUDA graphs. On the three cases whose selected mode is CUDA graphs, it excludes pruning alone in two, and neither has a pruning level meeting the budget. Because the compiled cases lack this measurement, Table~\ref{tab:screeningaudit} instead removes each affected stage entirely, counting prefill plus decode for autoregressive pruning. This weaker form excludes nothing on the grid. The weight-read screen excludes candidates in four cases, and each excluded candidate's own measured latency, after its pruning or quantization, is at least $2.67B$ (median $6.67B$).
This audit does not test how well the screens distinguish candidates close to the budget.
The weight-read bound assumes unchanged layer counts, at least 4-bit weights, and mandatory weight streaming from DRAM on each pass without inter-pass cache reuse. These assumptions must be checked before using the bound to exclude a deployment candidate.
Table~\ref{tab:screeningaudit} gives the counts for each search strategy.

\begin{table}[t]
\centering
\footnotesize
\caption{Comparison under a shared preference order. Except for Exhaustive, searches stop within each method family at its first qualifying pruning level or quantization group. Tested counts measurements; filtered counts excluded candidates; misses counts excluded budget-meeting candidates. Agreement counts cases returning the same choice as exhaustive enumeration, including cases with no qualifying option. The affected-stage row removes each affected stage entirely. The weight-read rule assumes Equation~\eqref{eq:weightread}. Counts exclude profiling and setup.}
\label{tab:screeningaudit}
\begin{tabular}{lrrrr}
\toprule
Method & tested & filtered & misses & agree \\
\midrule
Exhaustive & 156 & 0 & 0 & 13/13 \\
Stop when budget met & 107 & 0 & 0 & 13/13 \\
Affected-stage screen & 107 & 0 & 0 & 13/13 \\
Weight-read screen & 70 & 37 & 0 & 13/13 \\
Both screens & 70 & 37 & 0 & 13/13 \\
\bottomrule
\end{tabular}
\end{table}

\paragraph{Budget sensitivity.}
We repeat the budget check at different control periods for all sixteen Llama-based model--platform pairs, using the recorded measurements and the weight-read screen with datasheet bandwidth.
At a 33\,ms control period, the same 13 rows remain over their nominal budgets after mode selection, and the weight-read screen excludes 5 of them.
At the nominal 50\,ms period, the screen excludes 4 of the 13 over-budget rows.
At 100\,ms, 7 remain over budget and the screen excludes 2.
The OpenVLA A10G weight-read bound is approximately 39\,ms, so it does not exclude that row at either the nominal 50\,ms budget or the 47.5\,ms threshold after a 5\% reserve alone.
Using each configuration's p99 latency instead of its median changes none of these decisions. Across configurations using CUDA graphs or compilation, the median increase from median latency to p99 latency is 0.6\%.
Applying the deployment reserves of Section~\ref{sec:procedure}, preprocessing plus a 5\% margin, changes the selected setting in 4 cases.

\paragraph{Allowance for initial levels.}
In a replay of the recorded eager/CUDA-graph comparisons excluding LLaVA-VLA, each case fixes the model, device label, measurement campaign and control period. Software and power variants retain separate device labels. Of the 81 cases, 12 need pruning and have a budget-meeting level. We test initial-level allowances of 1.4\%, 3\%, 5.3\% and 10\%. Each retains a qualifying level in all 12 cases. The two larger allowances also select the preferred level in all 12; the smaller allowances do so in 10.
The allowance $\delta=5.3\%$ is a heuristic evaluated on this grid, not a bound for unseen settings.

\begin{table*}[t]
\centering
\footnotesize
\caption{Over-budget model--platform cases and the recorded settings, if any, that meet the inference-time budget before reserves. Task accuracy is not evaluated. Parentheses give milliseconds. Modes: g, graphs; c, reduce-overhead; a, max-autotune. Pruning is random. Choices follow Appendix~\ref{app:screeningobjective}'s preference order. The weight-read bound uses rated bandwidth. Bold exceeds $B$. A dash means no recorded option meets the budget, and n.m.\ means unmeasured in the row's mode. Recorded pruning levels vary by platform. The RTX 4090 records steps of 6.25\%.}
\label{tab:triage}
\renewcommand{\arraystretch}{0.9}
\setlength{\tabcolsep}{4pt}
\begin{adjustbox}{max width=\textwidth}
\begin{tabular}{@{}llcrrlll@{}}
\toprule
Model & GPU & mode & $T$ / $B$ & bound & pruning & quantisation & pair \\
\midrule
OpenVLA & H100 & a & 50.4 / 50 & 6 & 25\% (48) & n.m. & \na \\
OpenVLA & A10G & a & 242 / 50 & 39 & \na & \na & misses \\
OpenVLA & Thor & c & 419 / 50 & \textbf{86} & \na & \na & misses \\
OpenVLA & RTX 4090 & g & 142 / 50 & 23 & \na & n.m. & n.m. \\
OpenVLA-OFT & A10G & a & 159 / 150 & 7 & 25\% (142) & int8 compute (107) & \na \\
OpenVLA-OFT & Thor & c & 153 / 150 & 15 & 50\% (98) & int8 weights (121) & \na \\
LLaVA-VLA & H100 & c & 273 / 100 & 31 & \na & \na & misses \\
LLaVA-VLA & A10G & a & 1137 / 100 & \textbf{204} & \na & \na & misses \\
LLaVA-VLA & Thor & c & 2189 / 100 & \textbf{448} & \na & \na & misses \\
LLaVA-VLA & RTX 4090 & g & 695 / 100 & \textbf{121} & \na & n.m. & n.m. \\
CogACT & A10G & a & 97.5 / 50 & 6 & \na & \na & 75\% + int8 compute (49.6) \\
CogACT & Thor & c & 95.8 / 50 & 14 & \na & \na & misses \\
CogACT & RTX 4090 & g & 50.6 / 50 & 4 & 12.5\% (44) & n.m. & n.m. \\
\bottomrule
\end{tabular}
\end{adjustbox}
\end{table*}


\end{document}